\documentclass[sigconf]{acmart}

\AtBeginDocument{%
  \providecommand\BibTeX{{%
    \normalfont B\kern-0.5em{\scshape i\kern-0.25em b}\kern-0.8em\TeX}}}

\copyrightyear{2022}
\acmYear{2022}
\setcopyright{acmcopyright}
\acmConference[WWW '22]{Proceedings of the ACM Web Conference 2022}{April 25--29, 2022}{Virtual Event, Lyon, France}
\acmBooktitle{Proceedings of the ACM Web Conference 2022 (WWW '22), April 25--29, 2022, Virtual Event, Lyon, France}
\acmPrice{15.00}
\acmDOI{10.1145/3485447.3512242}
\acmISBN{978-1-4503-9096-5/22/04}

\usepackage{csvsimple}
\usepackage{datatool}
\usepackage{tabularx}
\usepackage{booktabs}
\usepackage{caption}
\usepackage{subcaption}
\usepackage{xstring}
\usepackage{multirow}
\usepackage{siunitx}

\usepackage[ruled,vlined]{algorithm2e}

\usepackage{soul} 
\usepackage[textsize=scriptsize,backgroundcolor=green,disable]{todonotes}

\begin{document}

\title{Measuring Annotator Agreement Generally across Complex Structured, Multi-object, and Free-text Annotation Tasks}


%
\author{Alexander Braylan}
\affiliation{%
  \institution{Dept.\ of Computer Science\\
  University of Texas at Austin \country{USA}}
}
\email{braylan@cs.utexas.edu}

\author{Omar Alonso}
\affiliation{%
  \institution{College of Computer Sciences\\Northeastern University \country{USA}}
}
\email{o.alonso@northeastern.edu}

\author{Matthew Lease}
\affiliation{%
  \institution{School of Information\\
  University of Texas at Austin \country{USA}}
}
\email{ml@utexas.edu}

\begin{abstract}

When annotators label data, a key metric for quality assurance is inter-annotator agreement (IAA): the extent to which annotators agree on their labels. Though many IAA measures exist for simple categorical and ordinal labeling tasks, relatively little work has considered more complex labeling tasks, such as structured, multi-object, and free-text annotations. Krippendorff's $\alpha$, best known for use with simpler labeling tasks, does have a distance-based formulation with broader applicability, but little work has studied its efficacy and consistency across complex annotation tasks.

We investigate the design and evaluation of IAA measures for complex annotation tasks, with evaluation spanning seven diverse tasks: image bounding boxes, image keypoints, text sequence tagging, ranked lists, free text translations, numeric vectors, and syntax trees.
We identify the difficulty of interpretability and the complexity of choosing a distance function as key obstacles in applying Krippendorff's $\alpha$ generally across these tasks. We propose two novel, more interpretable measures, showing they yield more consistent IAA measures across tasks and annotation distance functions. 

\end{abstract}



\maketitle


\section{Introduction}
\label{sec:intro}


Data annotations are often collected from human experts or crowdsourcing \cite{alonso2019practice} as part of the process for training and evaluating models. As an early and crucial node of the machine learning pipeline, it is important both to have quality labels \cite{aroyo22, northcutt2021pervasive} and to be able to measure label quality.
In real-world applications that require gathering many labels, measures of \textit{inter-annotator agreement} (IAA)  \cite{paun2022statistical} can be used to detect problems with data reliability stemming from task design, workers’ performance, or other causes \cite{alonso2013implementing}. Due to the wide variety of different annotation tasks, there is typically no single method for measuring agreement that is suitable for every purpose, and sometimes the inappropriate use and interpretation of such statistics can lead to wasted effort and resources or mis-specified and biased models.
A more comprehensive approach for this problem is to understand the complexity of the labeling task, identify an agreement metric that is explainable and fits the specific project requirements, in combination with other quality control mechanisms that can quantify the quality of a dataset.

In this paper, we investigate the use of IAA measures for ``complex'' annotation tasks \cite{braylan2020modeling,braylan2021aggregating} having large (finite or continuous) answer spaces, such as bounding boxes and keypoints in images, named entities in text, syntactic parse trees, free-text translations, ranked lists, and multi-dimensional numeric vectors. Most prior IAA studies assume relatively simple labeling tasks, such as classification or ordinal rating tasks. %


One of the most versatile IAA measures, 
Krippendorff's $\alpha$ \cite{krippendorff2004reliability}, can (in its most general form) be applied across diverse labeling tasks. As a baseline, we present empirical results for $\alpha$ across a variety of complex annotation tasks and task-specific distance functions for measuring annotation similarity.
However, we observe two important limitations of Krippendorff's $\alpha$.
First, $\alpha$ is difficult to interpret because its threshold for acceptable agreement varies greatly by task and distance function. This makes it confusing to understand when collected labels are of sufficient quality for use, especially with new tasks in which the task-specific $\alpha$ threshold is not yet known. 
Second, $\alpha$ requires selection of an appropriate distance function for the annotation task, and a poor choice can add noise, obscuring and underestimating agreement. This choice of distance function can be complicated as well. While prior work has relied on building simulators such as the Corpus Shuffle Tool (CST)~\cite{mathet2012manual} to evaluate distance functions, this requires creating an annotation simulator for each new labeling task of interest. 


Our innovation is to propose new,  distributional variants of Krippendorff's $\alpha$ that provide a conceptually and empirically more interpretable threshold for deciding that the data is ``good enough'', as well as clearer insight into selecting a distance function, without requiring either task-specific label noise simulation or gold data. 


\vspace{0.5em}
{\bf Contributions} of our work include:
\begin{itemize}
\vspace{-0.5em}

\item We provide a guide to the considerations and techniques for specifying an annotation distance function, that generalizes across various complex annotation tasks.
\item We identify two key limitations of the general form of Krippendorff’s $\alpha$ and provide novel alternatives.
\item A new IAA measure based on the Kolmogorov-Smirnov test \cite{massey1951kolmogorov} is shown to be particularly effective in evaluating distance functions for use with IAA, precluding need for either label noise simulation or gold data. 
\item A new IAA measure $\sigma$ provides a clear and task-general interpretation, with a lower bound to what fraction of observed label distances are significantly smaller than chance.
\item To support reproducibility, we share our code and data\footnote{\url{https://github.com/Praznat/annotationmodeling}}.
\end{itemize}

\section{Related Work}
\label{sec:related}

\subsection{Inter-annotator Agreement}
\label{sec:rw-iaa}

In collecting labeled data, it is useful to distinguish between {\em objective} tasks (in which a single best response is presumed to exist for each item) vs.\ {\em subjective} labeling tasks \cite{Nguyen16-hcomp} that expect diverse responses (e.g., soliciting personal preferences or opinions). Whereas high inter-annotator agreement (IAA)  \cite{paun2022statistical} is typically a goal with objective tasks, that is not the case with subjective tasks. 

Even with objective tasks, annotator disagreement can still be a useful signal to model training and evaluation \cite{aroyo2013crowd}, indicating varying confidence of ``ground truth'' labels for across items (e.g., due to corner-cases in annotation guidelines or difficult instances such as blurry images, etc.). Annotators may also cluster into different {\em schools of thought} \cite{tian2012learning} in interpreting or executing annotation guidelines due to different personal backgrounds, task ambiguity, etc. In addition, there are further risks of data bias \cite{sen2015turkers,mehrabi2021survey}: a pool of homogeneous annotators may agree with one another yet miss important problems with task guidelines or specific items that may be apparent to more diverse and representative annotators. 

The purpose of collecting multiple annotations per item is typically to assess and/or improve the quality of the labels, where the quality is presumed better when annotators agree (given the above caveats). Annotator agreement should not be confused with correctness; annotators can agree with one another yet share a systematic bias in collectively interpreting task guidelines differently than the author of those guidelines had intended. A common practice is thus to first check if annotators agree with one another (i.e., is the task clear?), then sample agreed-upon labels to ensure they are further consistent with what was actually desired from data collection. 

Given the inherent variability in human judgement as well as complexity of the collection process, disagreements can arise for a large variety of reasons, such as annotator heterogeneity (which is often desirable). Beyond demographics, annotators may vary in their training or skill, or in the effort they apply. Their labels may change over time from fatigue \cite{carterette2010effect} or calibration \cite{scholer2013effect}. 

Inter-annotator disagreement may also arise from heterogeneous items. Some items may be more difficult to annotate than other items \cite{whitehill2009whose}. Even the definition of difficulty itself can be divided into multiple types. For example, items might be more \textit{discriminating}, in that the more skilled annotators are much less prone to error than the less skilled ones, or more \textit{ambiguous} in that both skilled and less skilled annotators are equally prone to error \cite{baker2004item}.

\textbf{Global sources of inter-annotator disagreement} include a random noise factor that may affect any given annotation, as well as systematic problems in the annotation process such as an unreliable platform
or confusing instructions. These can stem from many ways in which annotations are collected, whether it be a crowdsourcing platform, an internal lab or team, managed workers, etc, with wide varieties in how tasks are 
designed and implemented.

Finally, one global source of measured inter-annotator disagreement is the method by which it is being measured.
Much of the prior work on measuring IAA is around improving these measures so that they do not show more or less agreement than what arises from the aforementioned factors. One of the major innovations from prior work is the \textit{chance correction}, or the separation of observed disagreements between annotations from what should be expected due to chance.
Many such chance-corrected agreement measures exist, including Scott's $\pi$ \cite{scott1955reliability}, Cohen's $\kappa$ \cite{cohen1960coefficient}, and Fleiss' $\kappa$ \cite{fleiss1971measuring}. The common approach for chance correction is to distinguish \textit{observed disagreements} $D_o$ from \textit{expected disagreements} $D_e$. Not performing such a correction can hinder the interpretability of the agreement measure, as the size of the possible and likely response spaces would heavily affect the magnitude of the measure.

Krippendorff's $\alpha$ \cite{krippendorff2004reliability} is a measure that aims to generalize many others. Not only can it handle any number of annotators and missing values, it also allows plugging in a \textit{distance function} that could in principle apply to any type of annotation for which such a function can be conceived. However, because of its design in the context of certain specific tasks such as content analysis, there is sometimes confusion around its definition.
When the literature refers to Krippendorff's $\alpha$, it may refer to either: i) the general formula $\alpha = 1 - \frac{\hat{D}_o}{\hat{D}_e}$ for computing agreement given a distance function $D(a,b)$; or ii) a specific distance function $D(a,b)$ to use in this general formula.

\citet{krippendorff2004reliability} prescribes distance functions for several kinds of data, including nominal, ordinal, interval, ratio, polar, and circular. Sometimes alternatives to Krippendorff's $\alpha$ are actually alternatives to the prescribed distance functions rather than the general form.
For example ``weighted Krippendorff's $\alpha$'' \cite{antoine2014weighted} distinguishes the use of a Euclidean distance function from a binary distance function, but relies on the same general form $1 - \frac{\hat{D}_o}{\hat{D}_e}$. In this paper, when we refer to Krippendorff's $\alpha$ we mean specifically the general formula and not any de facto prescriptions of distance functions. It is important to make this distinction in order to separate properties and criticisms of the general formula from properties and criticisms of specific distance functions that plug into it. This paper investigates challenges with both the general formula and with the variety and assessment of distance functions for complex annotations.
 
\vspace{-1em}
\subsection{Complex Annotations}
\label{sec:rw-ca}

\begin{figure}
\centering
  \fbox{\includegraphics[clip,width=\columnwidth]{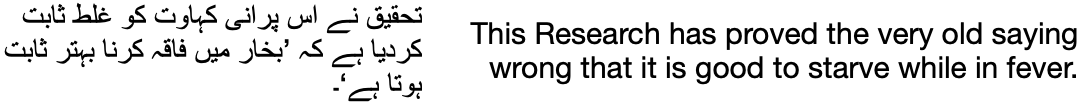}}
  \fbox{\includegraphics[clip,width=.8\columnwidth]{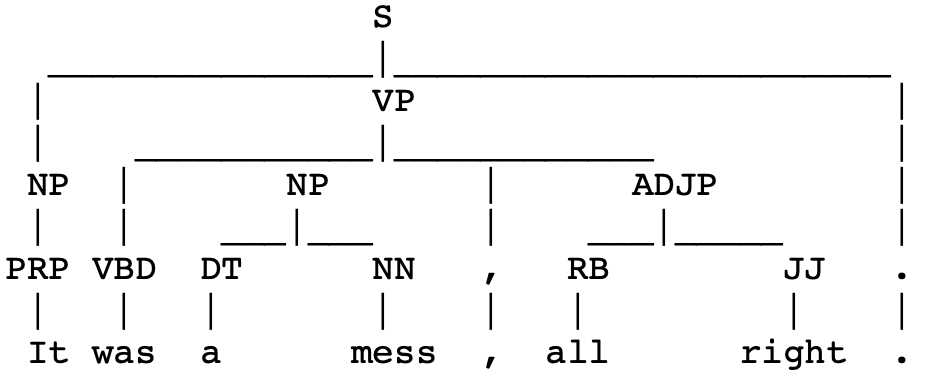}}
  \fbox{\includegraphics[clip,width=.8\columnwidth]{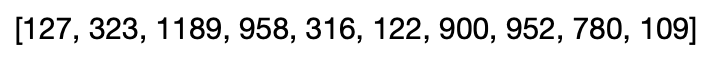}}

\includegraphics[clip,width=\columnwidth]{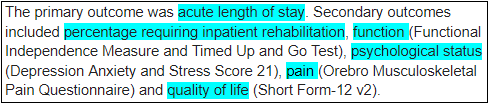}
\includegraphics[clip,width=(\columnwidth / 2 - 1pt)]{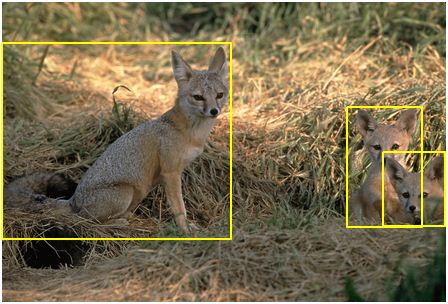}
\includegraphics[clip,width=(\columnwidth / 2 - 1pt)]{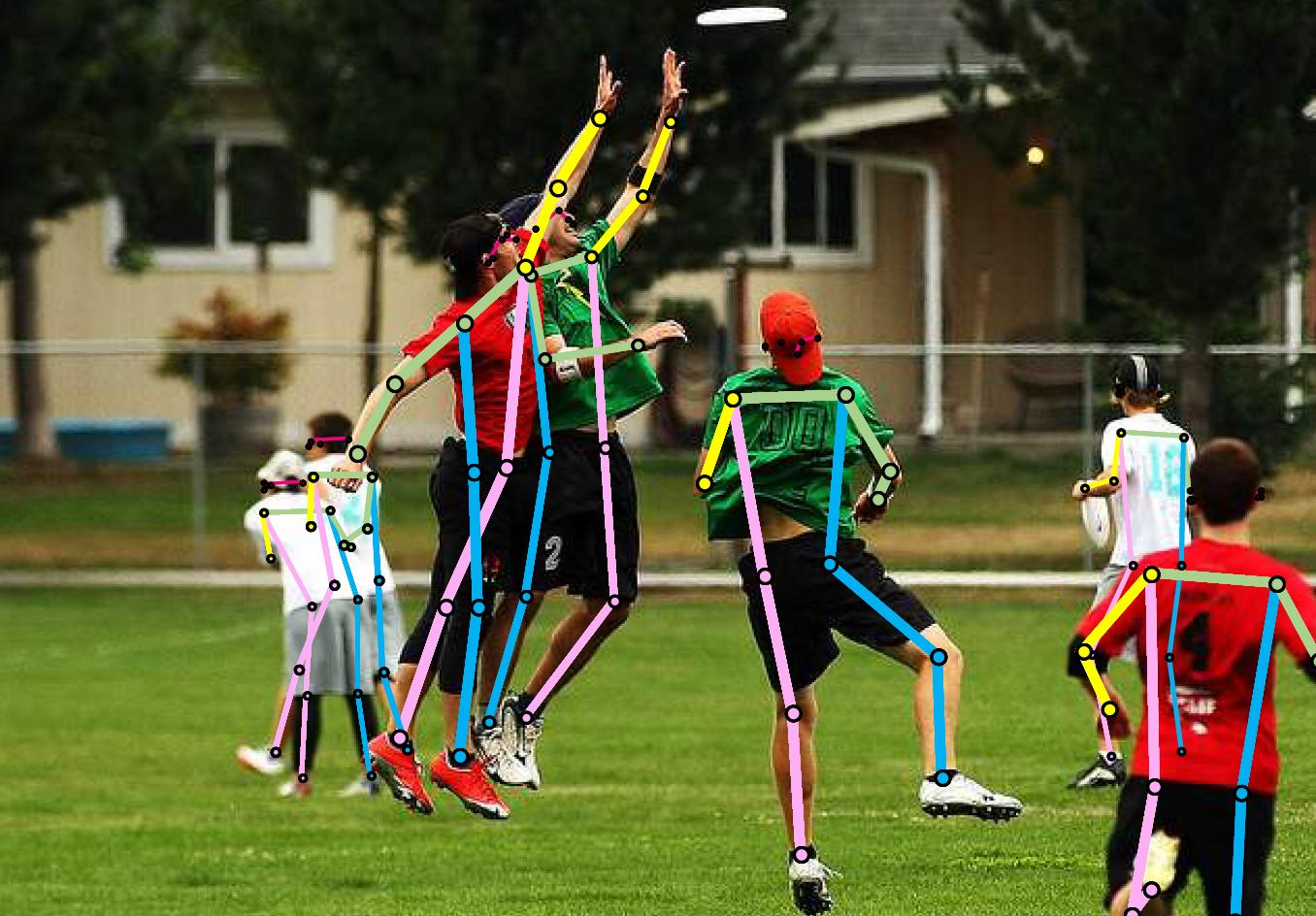}

\caption{Examples of complex annotations. Open-ended: language translation. Structured: a syntactic parse tree, and a ranked list of elements.  Multi-object: text sequences for information extraction, image bounding boxes for object detection, and image keypoints for pose estimation.}
\label{fig:examples}
\vspace{-2em}
\end{figure}

As practitioners seek to automate ever-more sophisticated tasks, annotation needed to train and evaluate models becomes increasingly complex. More complex tasks \cite{parameswaran2016optimizing,arous2020opencrowd,braylan2020modeling,braylan2021aggregating} may involve open-ended answer spaces (e.g., translation, transcription, extraction) or structured responses (e.g., annotating ranked lists, linguistic syntax or co-reference), such as those shown in Figure~\ref{fig:examples}. 

In addition, while simple labeling tasks like classification may require only a single label for each input {\em item} (e.g., a document or image), many complex annotation tasks require annotators to label an unknown number of {\em objects} per item — such as demarcating named-entities in a sentence \cite{sang2003introduction} or visual entities in an image \cite{branson2017lean}, as shown in the last three examples of Figure~\ref{fig:examples}.
\citet{mathet2015unified} refer to this part of a task as \textit{unitizing} -- defining the boundaries objects of interest -- in contrast to \textit{categorizing}, or assigning a class to each object.
\citet{braylan2020modeling} define \textit{complex annotations} as anything that is not a categorical or simple numerical label -- basically anything with a large to infinite response space. These tasks may also require  greater cognitive effort by annotators.

Recent work has explored modeling distances between annotations as a general approach to \textit{label aggregation} across diverse annotation types \cite{braylan2020modeling,braylan2021aggregating,li2020crowd,meir2020truth}.
A commonality among these is the use of a task-specific \textit{distance function} to convert each specific complex domain into a much more general numeric one. This is the same trick Krippendorff's $\alpha$ takes advantage of to handle complex annotations, but this prior work recommends the use of common \textit{evaluation functions} -- measures of a predicted label against gold -- which when inverted can be used as distance functions. An open question in this prior work is how to judge what distance function works best when there are multiple options.



\subsection{Agreement for Complex Annotations}
\label{sec:rw-iaaca}

While theoretically Krippendorff's $\alpha$ is applicable to any complex annotation task that has an available distance function, the prior work investigating such applications is sparse.
\citet{skjaerholt2014chance} investigates agreement metrics for dependency syntax trees, based on Krippendorff's $\alpha$. They evaluate different distance functions on syntax trees for use with Krippendorff's $\alpha$.
\citet{mathet2015unified} propose a \textit{Holistic $\gamma$} measure of IAA for various linguistic annotation tasks such as Named Entity Recognition and Discourse Framing.

Both of the above works depend on the use of a \textit{Corpus Shuffling Tool (CST)} \cite{mathet2012manual} for evaluating IAA measures.
 The insight of CST is to use a simulator of noise applied to complex annotations to control the expected amount of disagreement. A proposed measure of IAA is evaluated by plotting the amount of simulated error in a dataset against the measured agreement. An ideal measure should span from 0 to 1 as simulated error spans from 1 to 0, and this response should be strictly decreasing.
  Many of the conclusions from this prior work come from using CST to judge and rank various candidate distance functions to plug into Krippendorff's $\alpha$.
 
The downside of CST is that it is taxing to build simulators of possible error. Furthermore, these simulators might deviate from reality in the types of errors they capture. These reasons make it difficult to apply CST to a wide range of different complex annotation tasks, and therefore make it difficult to choose an appropriate distance function for a given task.

Choosing an effective distance function is crucial for measuring IAA. As discussed in Section~\ref{sec:rw-iaa}, a poor distance function can be a global contributor to total measured disagreement. That means the distance function competes as an explanation of disagreement with the other potential sources such as task difficulty, ambiguity, etc, that practitioners care about. The only way to isolate the effect of the distance function is to choose one that works better.


One desirable property of an IAA measure is thus its \textbf{absolute level to distinguish good distance functions from bad ones}. In Section~\ref{sec:alternative} we argue that Krippendorff's $\alpha$ does not fulfill this need, which is why prior work has relied on CST experiments.






\section{Generalizing to Complex Tasks}
\label{sec:generalizing}

In this section we lay out the assumptions, requirements, and procedure for measuring IAA in a way that generalizes across many diverse complex annotation modalities. 

The general formula for Krippendorff's $\alpha = 1 - \frac{\hat{D}_o}{\hat{D}_e}$, where $\hat{D}_o$ is the \textit{average distance observed}, and $\hat{D}_e$ is the \textit{average distance expected}. Observed distances $D_o$ are the set of \textit{within-item} pairwise distances between annotations, given a distance function $D$.

\vspace{-1em}
$$
D_o = \{D(a,b) \:|\: \textrm{ITEM}(a) = \textrm{ITEM}(b)\}
\hspace{7pt},\hspace{7pt} 
\hat{D}_o = \frac{1}{|D_o|}\sum_{d \in D_o} d
$$
\vspace{-1em}

The standard method for getting the set of expected
distances $D_e$ is to sample \textit{inter-item} pairwise distances between annotations, which is generally applicable to complex annotations as well.

\vspace{-1em}
$$
D_e = \{D(a,b) \:|\: \textrm{ITEM}(a) \neq \textrm{ITEM}(b)\}
\hspace{7pt},\hspace{7pt} 
\hat{D}_e = \frac{1}{|D_e|}\sum_{d \in D_e} d
$$
\vspace{-1em}

Recall that expected distances are used for chance correction. That means that expected distances should represent what a measured distance might be between randomly-made annotations. For example, taking translations from two different items (source sentences) is a reasonable proxy for randomly generating translations.

\subsection{Distance Function Properties}

This method for computing $D_e$ introduces one limitation: \textbf{distance functions must be applicable to pairs of annotations from different items}. As noted by \citet{skjaerholt2014chance}, such a limitation precludes a number of candidate distance functions for syntactic parse trees, including EVALB \cite{sekine1997evalb}, causing them to choose Tree Edit Distance (TED) which does not have this limitation.

Also noted in \citet{mathet2015unified} is that distance functions are \textit{metrics} that must fulfill the requirements of
\textbf{Non-negativity},
\textbf{Symmetry},
\textbf{Zero only for identical inputs}, and
\textbf{Triangle inequality}.
One may skip the triangle inequality requirement and call it a \textit{dissimilarity}, as in \citet{mathet2015unified}. However, we will continue to use the term ``distance'' in this paper, assuming the triangle inequality requirement without studying whether it is truly needed.

\subsection{Supporting Multi-object Items}

With multi-object labeling tasks (e.g., labeling named-entities in a text or bounding boxes in an image), annotators or prediction models must locate (aka {\em unitize} \cite{mathet2015unified}) and categorize 1-many objects in each input text/image. Evaluation metrics then score how well the set of annotated or predicted objects matches the true set of gold objects for the text/image. As \citet{braylan2021aggregating} note, evaluation metrics already exist for many such multi-object labeling tasks and often can be directly applied as distance functions.

Another general strategy for supporting multi-object labeling tasks is to induce a multi-object distance function $D_m(A,B)$ from a single-object distance function $D_s(a,b)$ by finding the minimum distance from each object in $A$ to each object in $B$ and vice-versa.
\vspace{-1em}
$$
\Delta(A,B) = \mathbb{E} \{ \text{min}(\{D_s(a, b) \mid b \in B \}) \mid a \in A \} \\
$$
\vspace{-2.5em}
$$
D_m(A,B) = \frac{\Delta(A,B) + \Delta(B,A)}{2}
$$
For example, \citet{mathet2015unified}
provide a comparable algorithm for computing an \textit{alignment} that minimizes local disagreements between units in different annotations. The benefit of this kind of approach is that it abstracts away everything except the choice of single-object distance function $D_s$, which can be easier to provide.

Yet another general approach for dealing with complex annotations is to first decompose them into simpler annotations and operate on these instead \cite{parameswaran2016optimizing,braylan2021aggregating}. For example, \citet{nguyen2017aggregating} relax traditionally ``strict'' NER scoring of exact spans \cite{sang2003introduction} by decomposing them into tokens and computing ``partial-credit'' scoring by token. Similarly, Jaccard index or Intersection-over-Union (IoU) decompose labeled image regions into pixels and measure partial-credit label distance based on area overlap. The benefit of decomposition is that since every annotator (implicitly) labels every low-level token/pixel, there is no need to align labels across annotators.   



Krippendorff's $\alpha$ plus an appropriate task-specific distance function 
provides a very general approach for measuring IAA across diverse types of complex annotations. However, questions remain as to how to choose between distance functions and how to interpret the IAA measure. In the following section we will discuss why Krippendorff's $\alpha$ presents difficulties towards these ends, and we propose methods to address those difficulties.




\begin{figure}[b]
\vspace{-1em}
\centering
\includegraphics[width=.85\columnwidth]{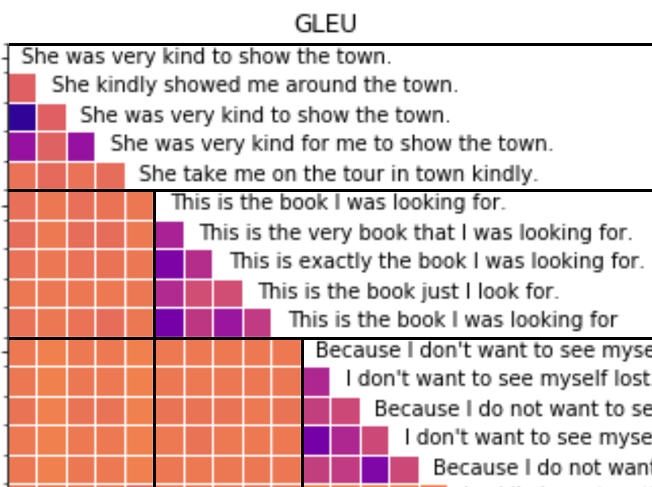}
\includegraphics[width=.8\columnwidth]{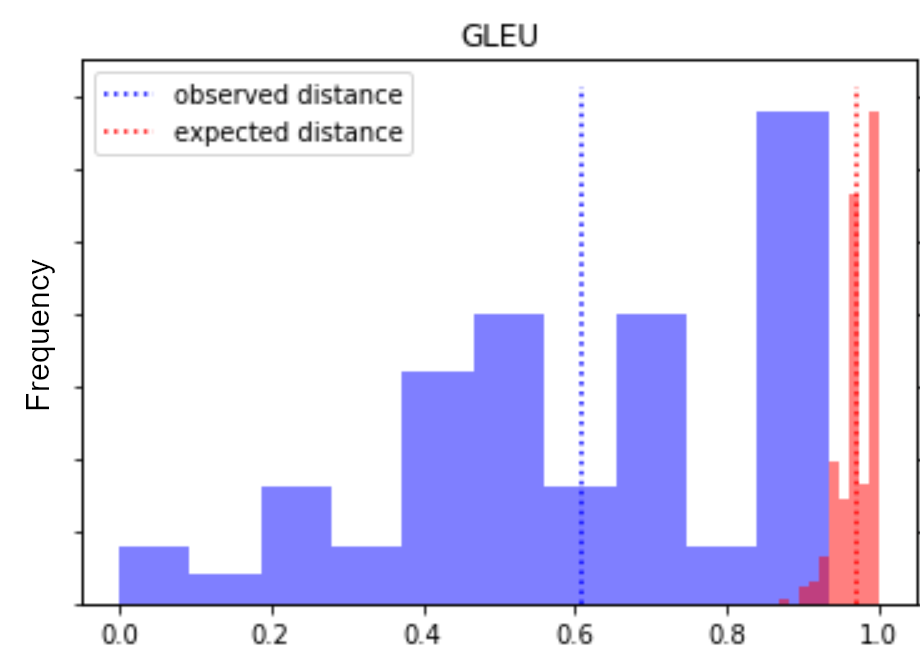}
\caption{\underline{Top}: Heatmap of GLEU distance between translations for three sentences (darker being more similar). High contrast between within-item distances $D_o$ and inter-item distances $D_e$ implies annotators are much more in agreement than random. \underline{Bottom}: Distribution of within-item GLEU distances $D_o$ and inter-item GLEU distances $D_e$ for same dataset. While the distribution for observed GLEU distances is quite wide, only a small portion of it overlaps the distribution of GLEU distances expected at random.}
\label{fig:gleu-dist}
\end{figure}

\section{Alternatives to Krippendorff's $\alpha$}
\label{sec:alternative}


So far we have assumed IAA should be measured for complex annotations using Krippendorff's $\alpha$, with the main question being what distance function to use. In this section, however, we argue that Krippendorff's $\alpha$ falls short when applied generally to complex annotations, for two reasons: its difficulty of interpretation and the complexity of using it to choose between distance functions. We provide two novel IAA measures to help address this.

\subsection{Problem of Interpretability}

IAA scores should aid interpetation of collected data. While a single agreement number does not provide all possible useful information about all possible sources of disagreement (such as annotator subjectivity, item ambiguity, etc.), this top level number should describe how much the overall data is \textbf{better than chance}. By looking at some examples we see that Krippendorff's $\alpha$ does not quite communicate that information sufficiently for complex data. 

For example, noting the contrast in Figure~\ref{fig:gleu-dist} between observed distances $D_o$ on the diagonal blocks and expected distances $D_e$ outside those, it is surprising that the calculated Krippendorff's $\alpha$ is only 0.35. This number seems low given the obvious contrast between within-item and inter-item distances and the overall qualitatively acceptable level of agreement between translations. \citet{krippendorff2004reliability} stipulated 0.667 as the ``lowest conceivable limit'' of acceptable $\alpha$, although anticipating that such guidelines would not likely extrapolate as far as something like translation data.
Still, one may ask what to do with such a low number in this example. Should we simply lower our expectations for acceptable $\alpha$ in Japanese-English translation tasks? What about other languages? What about other kinds of complex tasks?
This difficulty is what we mean when we discuss the \textit{interpretability} of an agreement measure. \textbf{To be \textit{interpretable}, a measure must have a stable notion across new annotation tasks of when annotations are “good enough” to proceed from pilot to production.}

The reason for Krippendorff's $\alpha$'s low interpretability across tasks stems from the fact that it compares an \textit{average} observed distance to an \textit{average} expected distance. And depending on the task, there may be a whole \textit{range} of acceptable distances that is lost when summarizing in the average. To clarify what we mean by this, consider the bottom of Figure~\ref{fig:gleu-dist}, in which the $D_o$ (in blue) and $D_e$ (in red) values are rearranged into histograms. While the distribution for observed distances is quite wide, only a small portion of it overlaps the distribution of distances expected at random.

\begin{figure}
\centering
\includegraphics[clip,width=\columnwidth]{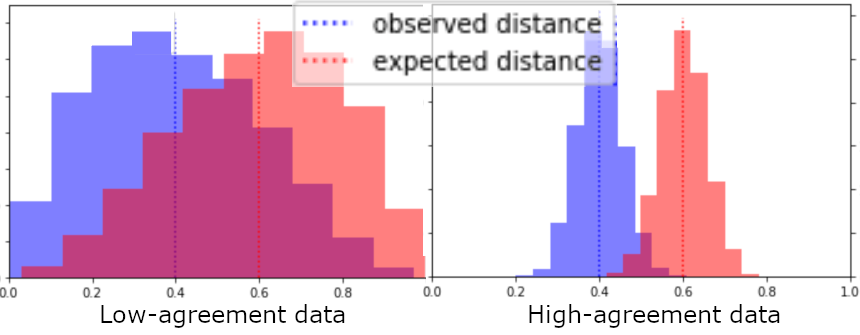}
\caption{Hypothetical datasets with low agreement (left) and high agreement (right). Purple-shaded regions denote area of overlap between observed annotation distances and distance expected from chance. Krippendorff's $\alpha$ is 0.33 for both datasets, whereas our measure $\sigma$ distinguishes 0.26 for the left and 0.98 for the right.}
\label{fig:low-high-agreement}
\vspace{-1em}
\end{figure}

To more generally see the shortcoming of comparing average $D_o$ and $D_e$, consider two hypothetical datasets illustrated in Figure~\ref{fig:low-high-agreement} which have the same average but different scales of $D_o$ and likewise for $D_e$. The dataset with the smaller scales intuitively has more agreement, as observed distances between annotations are more discernible from chance expected distances.
However, both datasets yield the same value for Krippendorff's $\alpha$.

\subsection{Problem of Distance Function Choice}

Section~\ref{sec:rw-iaaca} discussed the CST method used for judging and choosing between candidate distance functions for complex annotations. To reiterate, the CST method requires designing a simulator of errors specific to the complex annotation task at hand, which is impractical for the common practitioner who just wants a good way to measure agreement for their collected data. It would be ideal to have an easily computable single number that could grade a candidate distance function against another. Alas, Krippendorff's $\alpha$ cannot itself be used this way. As seen in the previous section and Figure~\ref{fig:low-high-agreement}, $\alpha$ is sometimes unable to differentiate a distance function that better separates $D_o$ and $D_e$ distributions from one that produces more overlap.
Furthermore, we will see in Section~\ref{sec:choosingdistance} that Krippendorff's $\alpha$ sometimes scores much higher for distance functions that we expect to be much worse, relative to other distance functions.


\subsection{Methods Proposed}

In order to better generalize to a wider set of tasks, we propose calculating agreement based on the difference in \textit{distribution} of $D_o$ and $D_e$ rather than the difference in their averages.

For comparing distributions there are a number of options. To grade options we consider two of the shortcomings of Krippendorff's $\alpha$ on complex annotations that we would like to remedy: the need for stable interpretability of the agreement score and the need for using the score to choose between distance functions. The former also depends on the latter -- agreement should be measured using the best distance function available in order to minimize the effect of the distance function as a source of disagreement.


One way to compare distributions is to first perform kernel density estimation to get a smooth probability distribution function (PDF) for the $D_e$ that can be evaluated for any individual observation from $D_o$. Then we can simply ask, ``{\em for each observed distance in $D_o$ what is the probability that a random draw from the distribution of expected disagreements $D_e$ could be smaller than it?}'' For any given observation $d$ from $D_o$, we can say that it is statistically significantly different from a random distance, if the cumulative distribution function (CDF, i.e. the integral of the PDF) of $D_e$ up to $d$ is smaller than $p=0.05$ (this $p$ threshold parameter is flexible but should be used consistently).
Finally, we note the fraction of $D_o$ deemed to have passed this one-sided significance test. This measure is easy to calculate and interpret: \textbf{the fraction of the observed distances that are unlikely to be drawn from random expected distances}. We denote this measure as $\sigma$:

\vspace{-1em}

$$
\sigma = \mathop{{}\mathbb{E}}_{d \in D_o} [~p > \int^{d} \textrm{PDF}(D_e)]
$$


A second option for comparing empirical distributions is the one-sided Kolmogorov-Smirnov (KS) test \cite{massey1951kolmogorov}, used to determine whether a sample ($D_o$) could be a drawn from a reference distribution ($D_e$). 
While $\sigma$ compares each observation from $D_o$ independently against the $D_e$ distribution, KS compares the whole sample.

\vspace{-1em}
$$
\textrm{KS} = \textrm{max}_x (\textrm{CDF}(D_o \leq x) - \textrm{CDF}(D_e \leq x))
$$


The measure of separation between distributions as a means to judge distance functions is supported by analogy to the \textit{metric learning} literature\cite{yang2006distance}, in which the objective is to learn a good metric: one that distinguishes data of the same class from data of different classes.
Another analogy is clustering, where the objective is to have the ratio of within-cluster distances to inter-cluster distances be as small as possible \cite{mehar2013determining}.
A key takeaway from these analogies is that {\bf a better distance function will better separate distributions of different classes}, without having to compare across different scenarios of simulated noise as CST does.
We will also show through experiments in Section~\ref{sec:choosingdistance} how our approach ranks distance functions appropriately according to how established literature would rank distance functions.

We calculate IAA KS measure as the complement of the statistical significance p-value ($1-p$) returned by the KS test. 
While we recommend the KS measure as a means to compare distance functions, we argue that the $\sigma$ measure has a more useful interpretation for deciding whether there is sufficient agreement in the data. Once a distance function has been chosen according to the highest KS score, we use the corresponding $\sigma$ for that distance function as a \textbf{lower bound} for how much of the data differs from chance. Just as ``indistinguishable from random'' does not necessarily mean random, the observed distances inside the expected distance distribution are not necessarily made in bad faith or due to errors or ambiguity. Therefore, a low $\sigma$ score does not necessarily mean the data should be discarded, it only means further investigation is necessary.

On the other hand, as a lower bound measure, a high $\sigma$ score is a good indication that there is not a significant amount of confusion or spamming or other causes for random-seeming annotations. A high $\sigma$ does not necessarily mean there is nothing left to investigate, but as a summary measure it serves the purpose of comparing agreements overall against chance.
What this does not guarantee is whether these annotations are useful enough to deploy in the real world. For that, it is still important to consider the needs and nuances that vary from task to task.
Overall we recommend interpreting IAA by using these proposed measures, but not exclusively as there can be other sources of data reliability issues that these measures do not specifically identify, such as bias.





\vspace{-1em}
\section{Experimental setup}

To compare inter-annotator agreement (IAA) metrics across a broad range of complex annotation types, we now summarize the datasets and distance functions used. 
For each dataset, we compare IAA using Krippendorff's $\alpha$ vs.\ our two new IAA metrics: KS and $\sigma$. 

\textbf{Vectors.} \underline{Dataset}. \citet{snow2008cheap} ask workers to score short text headlines according to six emotions on a [0-100] interval. Such data is typically modeled as a set of independent ordinal rating questions, neglecting that the same headline is being rated for all six emotions. In this work, we treat each headline as a single item to which the annotator assigns a complex, vector of six scalar values. \underline{Distance functions}. We compare coarse exact match (binary) vs.\ finer-grained Euclidean distance;  \citet{antoine2014weighted} recommend the latter for use of Krippendorff's $\alpha$ with ordinal annotations. 

\textbf{Translations.} \underline{Dataset}. \citet{li2019dataset}'s CrowdWSA2019 dataset of crowdsourced Japanese-to-English translations is drawn mostly from Japanese native speakers and non-native speakers of English. They encourage  beginner English speakers to participate and collect a dataset of more diverse quality than usually used to train machine translation models. \underline{Distance functions}. The four distance functions we compare are (in order of expected quality): Levenshtein, BLEU, GLEU, and BERTScore. Levenshtein is a general edit distance measure that ignores the finer nuances of natural language. BLEU \cite{papineni2002bleu} is a traditional baseline for evaluating translations. GLEU \cite{wu2016google} is a variant and improvement on BLEU, which is specialized for comparisons between individual sentences. BERTScore \cite{zhang2019bertscore} is the most modern approach, which takes advantages of the nuances of meaning and grammar baked into BERT embeddings. \citet{zhang2019bertscore} find it to correlate better with human judgements than quite a large assortment of competing measures. 

\textbf{Bounding Boxes.} \underline{Dataset}. \citet{braylan2021aggregating} share an image bounding box dataset in which each box is defined by an upper-left and lower-right vertex. An image may contain several visual entities to annotate, resulting in one to many bounding boxes per image to annotate. \underline{Distance functions}. We compare four distance functions: ``Count Diff'' measures only the difference in bounding box count between two annotations, L2 norm, Intersection Over Union (IoU), and Generalized IoU (GIoU). \citet{rezatofighi2019generalized} propose and show GIoU to be an improvement over standard IoU, which is itself an improvement over the L2 norm. 

\textbf{Named Entity Recognition.} \underline{Dataset}. \citet{sang2003introduction} share a NER dataset in which annotators highlight and categorize multiple spans of text within news articles. \underline{Distance functions}. We compare five distance functions varying in leniency. The coarsest ``Count Diff'' measures only the difference in named-entity count between two annotations. Leniency in the \textit{range} gives partial credit for range overlap, while leniency in the \textit{tag} simply ignores it. The strictest distance function requires both span and tag to be correct, while relaxations allow leniency in either or both span or tag.

\textbf{Keypoints.} \underline{Dataset}. \citet{braylan2021aggregating} share a synthetic dataset for image annotation using keypoints, generated by simulating various types of annotator noise over a base dataset from \citet{coco}. An image may contain multiple visual entities to be annotated with keypoints. \underline{Distance functions}. We compare three distance functions: the coarsest ``Count Diff'' of objects annotated, a coarse measurement of the IoU of the smallest boxes containing the keypoints, and the most commonly used function for comparing keypoints: Object Keypoint Similarity (OKS) \cite{ruggero2017benchmarking}.  

\textbf{Parse Trees.} \underline{Dataset}. \citet{braylan2020modeling} share a dataset of simulated syntactic parse data using the Brown corpus \cite{francis1979brown} and the Charniak parser \cite{mcclosky2006effective}. This dataset provides an alternate test for agreement measures on syntactic parse trees compared to  \citet{skjaerholt2014chance}'s syntactic parse data. Its source of simulated noise comes from error in sub-optimal machine parses, rather than random relabeling or reattaching of nodes. \underline{Distance functions}. Following \citet{skjaerholt2014chance}, we 
compare three variants of Tree Edit Distance (TED): $\alpha_{\textrm{plain}}$, $\alpha_{\textrm{norm}}$, and $\alpha_{\textrm{diff}}$, the latter two being different ways to normalize TED by the compared tree sizes. \citet{skjaerholt2014chance} finds $\alpha_{\textrm{plain}}$ to be the best distance function based on a CST analysis. 

\textbf{Ranked Lists.} \underline{Dataset}. \citet{braylan2020modeling} share a dataset of simulated ranked lists of elements. \underline{Distance functions}. We compare three distance functions: a coarse Kendall's $\tau$ over only the top-5 ranks, and $\tau$ and Spearman's $\rho$ calculated over the full ranking.

\begin{table}[]
\centering
\begin{tabular}{llrrr}
\toprule
Dataset & Distance f(x) &  $\alpha$ &  KS &  $\sigma$ \\
\midrule
Vector & Binary &        0.1277 &              0.5011 &          0.1151 \\
         & Euclidean \cite{antoine2014weighted}&        0.2146 &               0.5885 &          0.1593 \\
\midrule
Translations & Levenshtein &        0.2762 &          0.7735 &              0.5373 \\
         & BLEU \cite{papineni2002bleu} &        0.1816 &          0.8532 &              0.5791 \\
         & GLEU \cite{wu2016google} &        0.1656 &          0.8758 &              0.8100 \\
         & BERTScore \cite{zhang2019bertscore} &        0.4534 &          0.9085 &              0.8952 \\
\midrule
Bounding Boxes & Count Diff &        0.4365 &              0.6169 &          0.3736 \\
         & L2 &        0.6873 &              0.9130 &          0.7640 \\
         & IoU Score &        0.5046 &              0.9543 &          0.8418 \\
         & GIoU Score \cite{rezatofighi2019generalized} &        0.5069 &              0.9615 &          0.8711 \\
\midrule
NER & Count Diff &        0.3900 &              0.6205 &          0.1969 \\
        & both lenient &        0.4054 &              0.7816 &          0.6324 \\
         & both strict \cite{sang2003introduction} &        0.3324 &              0.7340 &          0.6620 \\
         & strict range &        0.3776 &              0.7688 &          0.6520 \\
         & strict tag &        0.3605 &               0.7848 &          0.6735 \\
\midrule
Keypoints & Count Diff &        0.0419 &              0.3871 &          0.4007 \\
         & IoU Score &        0.2924 &              0.7989 &          0.6278 \\
         & OKS Score &        0.6726 &              0.8715 &          0.5666 \\
\midrule
Parse Trees \cite{skjaerholt2014chance}         & $\alpha_{\textrm{diff}}$ &        0.8422 &              0.9815 &          0.9181 \\
& $\alpha_{\textrm{plain}}$ &        0.8768 &              0.9909 &          0.9601 \\
         & $\alpha_{\textrm{norm}}$ &        0.8626 &               0.9987 &          1.0000 \\
\midrule
Ranked Lists & Kendall's $\tau$ @5 &        0.2005 &              0.6099 &             0.6158 \\
        & Kendall's $\tau$ &        0.4915 &              0.9893 &             1.0000 \\
         & Spearman's $\rho$ &        0.5413 &               0.9867 &             1.0000 \\

\bottomrule
\end{tabular}
\caption{IAA metrics for different distance functions across datasets. Best $\alpha$ varies greatly between datasets, sometimes reaching very low levels despite these being mostly reliable datasets. For distinguishing between distance functions, $\alpha$ is also unreliable, making questionable preferences such as L2 > GIoU and Levenshtein > GLEU. Our measure KS's ordering of distance functions is more in line with expectations.}
\label{table:results}
\vspace{-2em}
\end{table}

\vspace{-.5em}

\section{Results}
\label{sec:evaluation}

We evaluate our methods with consideration of the following two objectives. First, the interpretation of our measures of IAA should be useful and general across a wide variety of different complex annotation tasks, with minimal need for domain-specific nuance. Second, our methods should help determine better distance functions, not only for measuring agreement but potentially also for aggregation and evaluation against gold.

\vspace{-1em}

\subsection{Score Interpretability}
\label{sec:interpretingresults}

Recall that an IAA score for a dataset should describe how much better the observed distances are from chance.
However, there are several methods for describing such difference from chance, and it can be difficult to compare these methods when there is no ground truth for how useful each one is.
One way to compare methods is to understand why they conflict with one another. Table~\ref{table:results} shows the different agreement scores for Krippendorff's $\alpha$, KS, and $\sigma$ across different distance functions for each dataset.

The top Krippendorff's $\alpha$ across distance functions varies drastically from dataset to dataset. This is expected because it is well known that $\alpha$ cannot be interpreted the same way across very different domains or across different distance functions \cite{artstein2008inter}. In order to get a ``good'' $\alpha$ score, $D_o$ and $D_e$ should cluster heavily around 0 and 1, respectively. For complex annotations, such results are very rare, as the distributions of $D_o$ and $D_e$ can be quite wide.

\begin{figure}[t]
  \centering
  \begin{subfigure}{.49\columnwidth}
    \centering
    \includegraphics[width=\linewidth]{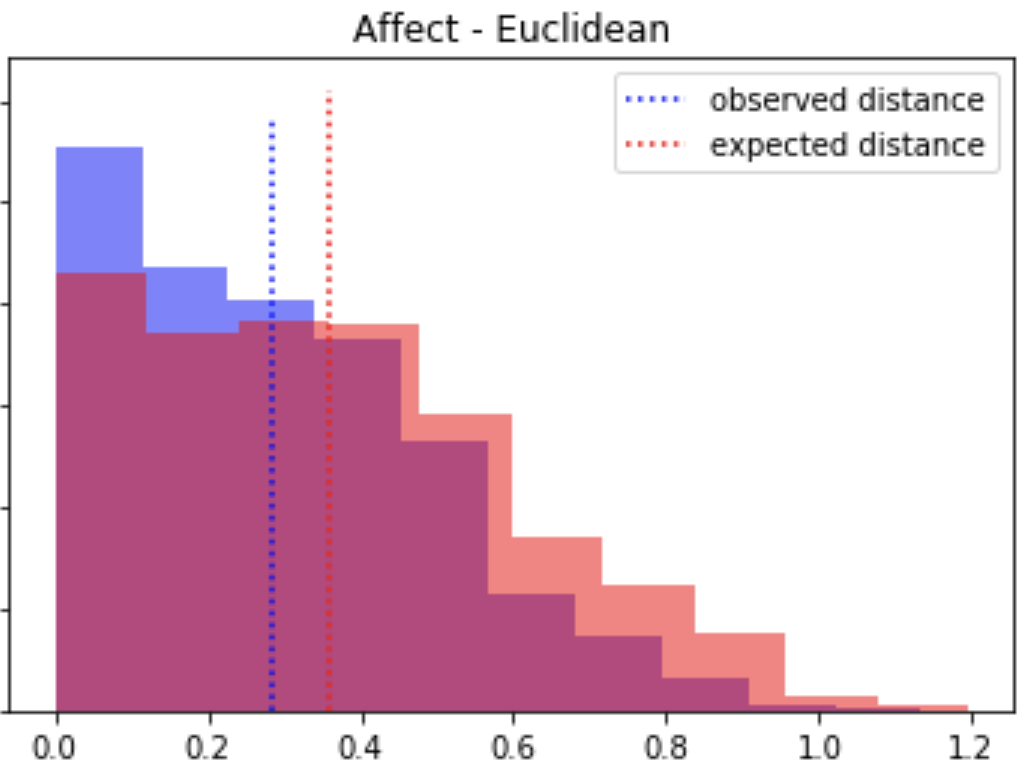}
    \caption{$D_e$ concentrated on the low end for vector interval labels.}
    \label{fig:affect_hist}
  \end{subfigure}%
  \hfill
  \begin{subfigure}{.49\columnwidth}
    \centering
    \includegraphics[width=\linewidth]{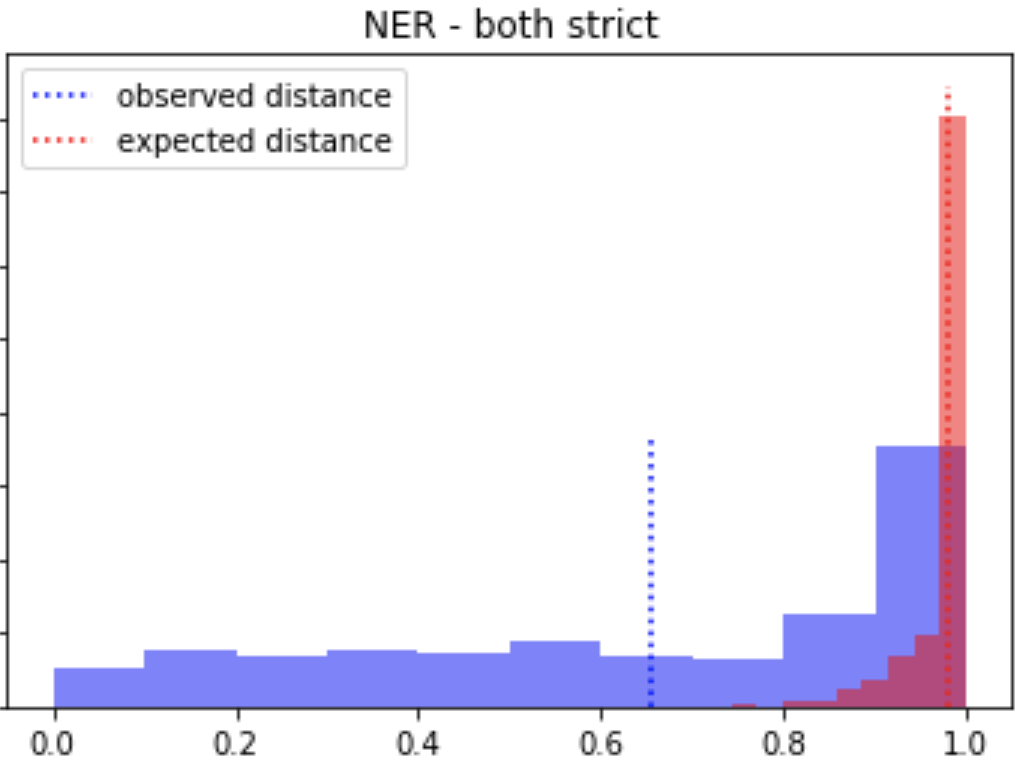}
    \caption{$D_o$ with a tall mode on the high end for NER exact-match.} 
    \label{fig:ner_hist}
  \end{subfigure}%
\vspace{-0.5em}
\caption{Examples of IAA underestimated by $\sigma$ due to a smaller response space than ideal for using $\sigma$. This can be remedied for NER by using a more lenient distance function.}
\vspace{-1.5em}
\end{figure}

On the other hand, KS seems to vary more from the choice in distance functions than from the task. The top KS score for each dataset is consistently high, with a single exceptionally low score for the Vector dataset. One explanation is that these are all fairly ``good'' datasets that were released publicly or generated with simulators. The entire sample of observed distances would need to be hard to differentiate from chance in order to produce a low KS score. Other than a serious problem in annotation task design, the only other ways KS can receive a very low score are if 1) a sub-optimal distance function obscures the signal separating the observed from the expected distributions, or 2) the space of possible responses is small enough to make the expected and observed distributions overlap significantly. Case 1) we discuss in Section~\ref{sec:choosingdistance} and is the main contributor to variation in KS for complex annotations with large response spaces. Case 2) seems to contribute to the low KS score for the Vector dataset, as Figure~\ref{fig:affect_hist} exemplifies how the observed and expected distances mostly overlap. Notably, most of the mass of the expected distance distribution falls on the low side, even more than would be expected for random independent draws from a six-dimensional uniform or Gaussian distribution, implying that the \textit{likely} annotation space across all items is much smaller than the \textit{possible} annotation space. This problem with interval data is pointed out by \citet{checco2017let}, whose alternative to Krippendorff's $\alpha$ is better suited to bounded numerical tasks like this one, though not applicable to more complex annotation types.


    

Our agreement measure $\sigma$ has the added value of interpretability over $\alpha$ and KS. A common struggle with $\alpha$ is that different distance functions result in very different  values \cite{artstein2008inter}, and the relative order of $\alpha$ between distance functions we see is difficult to explain as well. While KS yields a favorable ordering of distance functions, $\sigma$ actually has a relatively simple natural language explanation.

In Section~\ref{sec:alternative} we argue that the $\sigma$ score serves as a good lower bound measure of non-chance agreement. The reason it is a lower bound is that some of the real-life agreement can be occluded and underestimated due to the choice of distance function or nature of the annotation task.
For example, using the Count Diff distance for bounding boxes, we see that the $\sigma$ of 0.3736 only considers agreement on the number of boxes and ignores the tendency for annotators to agree on their boundaries.
Diagnosing a low $\sigma$ more generally can involve looking at histograms of the $D_o$ and $D_e$ distributions.
Examining the behavior of $\sigma$ across various datasets and distance functions, we find some common pathologies: 

\noindent~~$\blacktriangleright$ \textbf{Mode of expected distances near the low boundary.} 
As seen in the example of numeric vectors, this case can be common for lower-dimensional interval spaces. As $\sigma$ is developed for more complex annotation tasks with large response spaces, something like \citet{checco2017let}'s $\Phi$ may be more appropriate for interval annotations. This is also seen for certain distance functions that greatly constrain the response space, such as Count Diff.

\noindent~~$\blacktriangleright$ \textbf{Mode of observed distances near the high boundary.} 
This case can occur when there is a common cause of zero-credit comparison between annotations. For example, in the NER task, the stricter the distance function (e.g. ranges and tags must match exactly) the greater the density at the high-boundary mode becomes for observed distances (see Figure~\ref{fig:ner_hist}). 

\noindent~~$\blacktriangleright$ \textbf{Multi-modal observed or expected distances.} 
Both of the previous two cases can also be instances where the distance distributions exhibit multiple modes. A general concern is whether the complex response space is actually contracted in some way due to distance function or task design. For example, the task may have a first step asking whether an object exists in an image (binary response), followed by an image annotation if it exists at all. This example could result in a very tall mode for the instances where the first step precludes any following complex annotation.

Absent these problems, a low $\sigma$ score might just mean low agreement. For example, the Keypoints dataset is simulated with a large amount of noise in the location, rotation, and magnitude of the keypoint annotations, perhaps justifying its low $\sigma=0.5666$.


\vspace{-1em} 
\subsection{Comparing Distance Functions}
\label{sec:choosingdistance}

To assess how well our KS approach for choosing a distance function works, we compare the KS rankings of distance functions against ``expected'' orderings. The intuition behind these expected orderings stems from two lines of prior work. First, for each specific task or domain, prior work has often established current state-of-the-art distance functions (or evaluation metrics) for each task.
Second, prior CST work \cite{mathet2012manual} also induced an ordering of distance functions across different tasks, showing that that weaker distance functions were less sensitive to detecting label error, with less correlation observed between annotator agreement and simulated errors. During development we also conducted CST experiments and confirmed (unsurprisingly) that our agreement measures correlate negatively with increasing injected noise for stronger distance functions (Figure~\ref{fig:ner-cst} in Appendix).
Furthermore, our KS approach yields consistent distance function orderings with prior work and without requiring a task-specific noise simulator, which is particularly useful for novel annotation tasks lacking prior work.
One interesting finding is that ``fine-grained'' distance functions that capture more meaningful information than ``coarse'' ones tend to perform better.


We now discuss the resulting IAA scores seen in Table~\ref{table:results}.

\textbf{Numeric Vector.} All three measures show the finer-grained Euclidean distance outperforming mean element-wise binary exact match, consistent with \citet{antoine2014weighted}'s recommendation of Euclidean distance over binary for use on ordinal annotations. 

\textbf{Translations.} 
Both of our measures yield the expected order of distance functions (from best to worst): BERTScore, GLEU, BLEU, Levenshtein. Note that using Krippendorff's $\alpha$ here to compare the means rather than the full distributions of expected and observed differences ranks Levenshtein  above BLEU and GLEU.

\textbf{Bounding boxes.} 
Krippendorff's $\alpha$ is actually highest by far for L2, again indicating its inappropriateness for comparing distance functions simply in terms of levels of disagreement. Both of our measures on the other hand yield the expected order of distance functions (from best to worst): GIoU, IoU, L2, Count Diff.

\textbf{Named Entity Recognition.}
As expected, the coarsest ``Count Diff'' distance function is the least discriminating. We see lenient distance functions slightly outperform stricter ones, 
particularly leniency in the range, according to KS. Whereas leniency in range creates finer-granularity partial-credit in measuring distance, leniency in tag (i.e., ignoring tags) actually makes the distance measure coarser since categorical tags do not have any obvious notion of ``nearby'' for awarding partial credit, and ignoring tags entirely makes the distance function less discriminating. While in prior work, models trained on this dataset were evaluated using a strict metric \cite{sang2003introduction}, a more lenient metric that gives partial credit for range overlap provides finer-granularity of distance for calculating IAA. Our KS findings are consistent with CST findings \cite{mathet2015unified} on other datasets:  finer-grained distance measures beat coarse, binary ones.

\textbf{Keypoints.} 
The order of distance functions under KS from best to worst matches expectations: OKS, IoU, and Count Diff.

\textbf{Parse Trees.} 
Compared to \citet{skjaerholt2014chance}, our dataset shows similar KS and $\sigma$ scores across all three distance functions. The KS score for $\alpha_{\textrm{diff}}$ is slightly worse than the others, also consistent. 

\textbf{Ranked lists.} As expected, the coarser measure of Kendall's $\tau$ over only the top-5 ranks performs worst. Over the full ranking, both distance functions yield similar values for all agreement metrics. This is unsurprising because both correlation functions tend to return similar p-values for the same given data 
\cite{myers2013research}.





\vspace{-1.em} 
\section{Conclusion}

Because human annotations are pivotal in training and testing machine learning systems, it is important to have both reliable labels and effective ways to assess label quality. This is challenging due to the many possible sources of label disagreement and great variation in the nature of annotation across different labeling tasks, especially with ``complex'' labeling tasks \cite{braylan2020modeling,braylan2021aggregating} having large (finite or continuous) answer spaces. A common approach, inter-annotator agreement (IAA), supports assessment label quality on the basis of agreement between annotators, without assumption of any oracle ``gold standard''. However, most IAA methods do not generalize to complex labeling tasks. While the most general (and less known) form of Krippendorff's $\alpha$ \cite{krippendorff2004reliability} can be used, we showed two key limitations of it: difficulty identifying suitable distance functions and interpreting $\alpha$ across tasks and distance functions.

To address this, we described two novel IAA measures that offer greater conceptual and empirical interpretability than $\alpha$
for assessing when human annotations for complex labeling tasks are ``good enough'' to be used. 
Empirical testing across seven diverse complex annotation tasks shows how these measures add great value toward assessing IAA for complex annotations.

Various limitations remain for future work. For example, once we have isolated as best as possible the \textit{global} sources of disagreement from noise and distance function, how do we go further in diagnosing the contributions from annotator and item heterogeneity, without which we cannot fully understand IAA? How do we use IAA to predict how useful annotations will be after aggregation?

{\small 
\textbf{Acknowledgements}. We thank the many talented Amazon Mechanical Turk workers who contributed to our study. This research was supported in part by the Knight Foundation, the Micron Foundation, and Good Systems (\url{https://goodsystems.utexas.edu}), 
a UT Austin Grand Challenge to develop responsible AI technologies. Our opinions are entirely our own.
}

\bibliographystyle{ACM-Reference-Format}
\bibliography{www22.bib}


\begin{thebibliography}{48}


\ifx \showCODEN    \undefined \def \showCODEN     #1{\unskip}     \fi
\ifx \showDOI      \undefined \def \showDOI       #1{#1}\fi
\ifx \showISBNx    \undefined \def \showISBNx     #1{\unskip}     \fi
\ifx \showISBNxiii \undefined \def \showISBNxiii  #1{\unskip}     \fi
\ifx \showISSN     \undefined \def \showISSN      #1{\unskip}     \fi
\ifx \showLCCN     \undefined \def \showLCCN      #1{\unskip}     \fi
\ifx \shownote     \undefined \def \shownote      #1{#1}          \fi
\ifx \showarticletitle \undefined \def \showarticletitle #1{#1}   \fi
\ifx \showURL      \undefined \def \showURL       {\relax}        \fi
\providecommand\bibfield[2]{#2}
\providecommand\bibinfo[2]{#2}
\providecommand\natexlab[1]{#1}
\providecommand\showeprint[2][]{arXiv:#2}

\bibitem[\protect\citeauthoryear{Alonso}{Alonso}{2013}]%
        {alonso2013implementing}
\bibfield{author}{\bibinfo{person}{Omar Alonso}.}
  \bibinfo{year}{2013}\natexlab{}.
\newblock \showarticletitle{Implementing crowdsourcing-based relevance
  experimentation: an industrial perspective}.
\newblock \bibinfo{journal}{\emph{Information retrieval}} \bibinfo{volume}{16},
  \bibinfo{number}{2} (\bibinfo{year}{2013}), \bibinfo{pages}{101--120}.
\newblock


\bibitem[\protect\citeauthoryear{Alonso}{Alonso}{2019}]%
        {alonso2019practice}
\bibfield{author}{\bibinfo{person}{Omar Alonso}.}
  \bibinfo{year}{2019}\natexlab{}.
\newblock \showarticletitle{The practice of crowdsourcing}.
\newblock \bibinfo{journal}{\emph{Synthesis Lectures on Information Concepts,
  Retrieval, and Services}} \bibinfo{volume}{11}, \bibinfo{number}{1}
  (\bibinfo{year}{2019}), \bibinfo{pages}{1--149}.
\newblock


\bibitem[\protect\citeauthoryear{Antoine, Villaneau, and Lefeuvre}{Antoine
  et~al\mbox{.}}{2014}]%
        {antoine2014weighted}
\bibfield{author}{\bibinfo{person}{Jean-Yves Antoine}, \bibinfo{person}{Jeanne
  Villaneau}, {and} \bibinfo{person}{Ana{\"\i}s Lefeuvre}.}
  \bibinfo{year}{2014}\natexlab{}.
\newblock \showarticletitle{Weighted Krippendorff's alpha is a more reliable
  metrics for multi-coders ordinal annotations: experimental studies on
  emotion, opinion and coreference annotation.}. In
  \bibinfo{booktitle}{\emph{EACL 2014}}. \bibinfo{pages}{10--p}.
\newblock


\bibitem[\protect\citeauthoryear{Arous, Yang, Khayati, and
  Cudr{\'e}-Mauroux}{Arous et~al\mbox{.}}{2020}]%
        {arous2020opencrowd}
\bibfield{author}{\bibinfo{person}{Ines Arous}, \bibinfo{person}{Jie Yang},
  \bibinfo{person}{Mourad Khayati}, {and} \bibinfo{person}{Philippe
  Cudr{\'e}-Mauroux}.} \bibinfo{year}{2020}\natexlab{}.
\newblock \showarticletitle{Opencrowd: A human-ai collaborative approach for
  finding social influencers via open-ended answers aggregation}. In
  \bibinfo{booktitle}{\emph{Proceedings of The Web Conference 2020}}.
  \bibinfo{pages}{1851--1862}.
\newblock


\bibitem[\protect\citeauthoryear{Aroyo, Lease, Paritosh, and
  Schaekermann}{Aroyo et~al\mbox{.}}{2022}]%
        {aroyo22}
\bibfield{author}{\bibinfo{person}{Lora Aroyo}, \bibinfo{person}{Matthew
  Lease}, \bibinfo{person}{Praveen Paritosh}, {and} \bibinfo{person}{Mike
  Schaekermann}.} \bibinfo{year}{2022}\natexlab{}.
\newblock \showarticletitle{{Data Excellence for AI: Why Should You Care}}.
\newblock \bibinfo{journal}{\emph{ACM Interactions}} \bibinfo{volume}{29},
  \bibinfo{number}{2} (\bibinfo{year}{2022}).
\newblock
\newblock
\shownote{March-April.}


\bibitem[\protect\citeauthoryear{Aroyo and Welty}{Aroyo and Welty}{2013}]%
        {aroyo2013crowd}
\bibfield{author}{\bibinfo{person}{Lora Aroyo} {and} \bibinfo{person}{Chris
  Welty}.} \bibinfo{year}{2013}\natexlab{}.
\newblock \showarticletitle{Crowd truth: Harnessing disagreement in
  crowdsourcing a relation extraction gold standard}.
\newblock \bibinfo{journal}{\emph{WebSci2013. ACM}} \bibinfo{volume}{2013},
  \bibinfo{number}{2013} (\bibinfo{year}{2013}).
\newblock


\bibitem[\protect\citeauthoryear{Artstein and Poesio}{Artstein and
  Poesio}{2008}]%
        {artstein2008inter}
\bibfield{author}{\bibinfo{person}{Ron Artstein} {and} \bibinfo{person}{Massimo
  Poesio}.} \bibinfo{year}{2008}\natexlab{}.
\newblock \showarticletitle{Inter-coder agreement for computational
  linguistics}.
\newblock \bibinfo{journal}{\emph{Computational Linguistics}}
  \bibinfo{volume}{34}, \bibinfo{number}{4} (\bibinfo{year}{2008}),
  \bibinfo{pages}{555--596}.
\newblock


\bibitem[\protect\citeauthoryear{Baker and Kim}{Baker and Kim}{2004}]%
        {baker2004item}
\bibfield{author}{\bibinfo{person}{Frank~B Baker} {and}
  \bibinfo{person}{Seock-Ho Kim}.} \bibinfo{year}{2004}\natexlab{}.
\newblock \bibinfo{booktitle}{\emph{Item response theory: Parameter estimation
  techniques}}.
\newblock \bibinfo{publisher}{CRC press}.
\newblock


\bibitem[\protect\citeauthoryear{Branson, Van~Horn, and Perona}{Branson
  et~al\mbox{.}}{2017}]%
        {branson2017lean}
\bibfield{author}{\bibinfo{person}{Steve Branson}, \bibinfo{person}{Grant
  Van~Horn}, {and} \bibinfo{person}{Pietro Perona}.}
  \bibinfo{year}{2017}\natexlab{}.
\newblock \showarticletitle{Lean crowdsourcing: Combining humans and machines
  in an online system}. In \bibinfo{booktitle}{\emph{Proceedings of the IEEE
  Conference on Computer Vision and Pattern Recognition}}.
  \bibinfo{pages}{7474--7483}.
\newblock


\bibitem[\protect\citeauthoryear{Braylan and Lease}{Braylan and Lease}{2020}]%
        {braylan2020modeling}
\bibfield{author}{\bibinfo{person}{Alexander Braylan} {and}
  \bibinfo{person}{Matthew Lease}.} \bibinfo{year}{2020}\natexlab{}.
\newblock \showarticletitle{Modeling and Aggregation of Complex Annotations via
  Annotation Distances}. In \bibinfo{booktitle}{\emph{Proceedings of The Web
  Conference 2020}}. \bibinfo{pages}{1807--1818}.
\newblock


\bibitem[\protect\citeauthoryear{Braylan and Lease}{Braylan and Lease}{2021}]%
        {braylan2021aggregating}
\bibfield{author}{\bibinfo{person}{Alexander Braylan} {and}
  \bibinfo{person}{Matthew Lease}.} \bibinfo{year}{2021}\natexlab{}.
\newblock \showarticletitle{Aggregating Complex Annotations via Merging and
  Matching}. In \bibinfo{booktitle}{\emph{Proceedings of the 27th ACM SIGKDD
  Conference on Knowledge Discovery \& Data Mining}}. \bibinfo{pages}{86--94}.
\newblock


\bibitem[\protect\citeauthoryear{Carterette and Soboroff}{Carterette and
  Soboroff}{2010}]%
        {carterette2010effect}
\bibfield{author}{\bibinfo{person}{Ben Carterette} {and} \bibinfo{person}{Ian
  Soboroff}.} \bibinfo{year}{2010}\natexlab{}.
\newblock \showarticletitle{The effect of assessor error on IR system
  evaluation}. In \bibinfo{booktitle}{\emph{Proceedings of the 33rd
  international ACM SIGIR conference on Research and development in information
  retrieval}}. \bibinfo{pages}{539--546}.
\newblock


\bibitem[\protect\citeauthoryear{Checco, Roitero, Maddalena, Mizzaro, and
  Demartini}{Checco et~al\mbox{.}}{2017}]%
        {checco2017let}
\bibfield{author}{\bibinfo{person}{Alessandro Checco}, \bibinfo{person}{Kevin
  Roitero}, \bibinfo{person}{Eddy Maddalena}, \bibinfo{person}{Stefano
  Mizzaro}, {and} \bibinfo{person}{Gianluca Demartini}.}
  \bibinfo{year}{2017}\natexlab{}.
\newblock \showarticletitle{Let's agree to disagree: Fixing agreement measures
  for crowdsourcing}. In \bibinfo{booktitle}{\emph{Fifth AAAI Conference on
  Human Computation and Crowdsourcing}}.
\newblock


\bibitem[\protect\citeauthoryear{COCO}{COCO}{2020}]%
        {coco}
\bibfield{author}{\bibinfo{person}{COCO}.} \bibinfo{year}{2020}\natexlab{}.
\newblock \bibinfo{title}{Common Objects in Context (COCO)}.
\newblock
\newblock
\urldef\tempurl%
\url{https://cocodataset.org/}
\showURL{%
\tempurl}
\newblock
\shownote{Accessed: 2021-10-18.}


\bibitem[\protect\citeauthoryear{Cohen}{Cohen}{1960}]%
        {cohen1960coefficient}
\bibfield{author}{\bibinfo{person}{Jacob Cohen}.}
  \bibinfo{year}{1960}\natexlab{}.
\newblock \showarticletitle{A coefficient of agreement for nominal scales}.
\newblock \bibinfo{journal}{\emph{Educational and psychological measurement}}
  \bibinfo{volume}{20}, \bibinfo{number}{1} (\bibinfo{year}{1960}),
  \bibinfo{pages}{37--46}.
\newblock


\bibitem[\protect\citeauthoryear{Fleiss}{Fleiss}{1971}]%
        {fleiss1971measuring}
\bibfield{author}{\bibinfo{person}{Joseph~L Fleiss}.}
  \bibinfo{year}{1971}\natexlab{}.
\newblock \showarticletitle{Measuring nominal scale agreement among many
  raters.}
\newblock \bibinfo{journal}{\emph{Psychological bulletin}}
  \bibinfo{volume}{76}, \bibinfo{number}{5} (\bibinfo{year}{1971}),
  \bibinfo{pages}{378}.
\newblock


\bibitem[\protect\citeauthoryear{Francis and Kucera}{Francis and
  Kucera}{1979}]%
        {francis1979brown}
\bibfield{author}{\bibinfo{person}{W~Nelson Francis} {and}
  \bibinfo{person}{Henry Kucera}.} \bibinfo{year}{1979}\natexlab{}.
\newblock \showarticletitle{Brown corpus manual}.
\newblock \bibinfo{journal}{\emph{Letters to the Editor}} \bibinfo{volume}{5},
  \bibinfo{number}{2} (\bibinfo{year}{1979}), \bibinfo{pages}{7}.
\newblock


\bibitem[\protect\citeauthoryear{Krippendorff}{Krippendorff}{2004}]%
        {krippendorff2004reliability}
\bibfield{author}{\bibinfo{person}{Klaus Krippendorff}.}
  \bibinfo{year}{2004}\natexlab{}.
\newblock \showarticletitle{Reliability in content analysis: Some common
  misconceptions and recommendations}.
\newblock \bibinfo{journal}{\emph{Human communication research}}
  \bibinfo{volume}{30}, \bibinfo{number}{3} (\bibinfo{year}{2004}),
  \bibinfo{pages}{411--433}.
\newblock


\bibitem[\protect\citeauthoryear{Li}{Li}{2020}]%
        {li2020crowd}
\bibfield{author}{\bibinfo{person}{Jiyi Li}.} \bibinfo{year}{2020}\natexlab{}.
\newblock \showarticletitle{Crowdsourced Text Sequence Aggregation Based on
  Hybrid Reliability and Representation}. In
  \bibinfo{booktitle}{\emph{Proceedings of the 43rd International ACM SIGIR
  Conference on Research and Development in Information Retrieval}} (Virtual
  Event, China) \emph{(\bibinfo{series}{SIGIR '20})}.
  \bibinfo{publisher}{Association for Computing Machinery},
  \bibinfo{address}{New York, NY, USA}, \bibinfo{pages}{1761–1764}.
\newblock
\showISBNx{9781450380164}
\urldef\tempurl%
\url{https://doi.org/10.1145/3397271.3401239}
\showDOI{\tempurl}


\bibitem[\protect\citeauthoryear{Li and Fukumoto}{Li and Fukumoto}{2019}]%
        {li2019dataset}
\bibfield{author}{\bibinfo{person}{Jiyi Li} {and} \bibinfo{person}{Fumiyo
  Fukumoto}.} \bibinfo{year}{2019}\natexlab{}.
\newblock \showarticletitle{A Dataset of Crowdsourced Word Sequences:
  Collections and Answer Aggregation for Ground Truth Creation}. In
  \bibinfo{booktitle}{\emph{Proceedings of the First Workshop on Aggregating
  and Analysing Crowdsourced Annotations for NLP}}. \bibinfo{pages}{24--28}.
\newblock


\bibitem[\protect\citeauthoryear{Massey~Jr}{Massey~Jr}{1951}]%
        {massey1951kolmogorov}
\bibfield{author}{\bibinfo{person}{Frank~J Massey~Jr}.}
  \bibinfo{year}{1951}\natexlab{}.
\newblock \showarticletitle{The Kolmogorov-Smirnov test for goodness of fit}.
\newblock \bibinfo{journal}{\emph{Journal of the American statistical
  Association}} \bibinfo{volume}{46}, \bibinfo{number}{253}
  (\bibinfo{year}{1951}), \bibinfo{pages}{68--78}.
\newblock


\bibitem[\protect\citeauthoryear{Mathet, Widl{\"o}cher, Fort, Fran{\c{c}}ois,
  Galibert, Grouin, Kahn, Rosset, and Zweigenbaum}{Mathet
  et~al\mbox{.}}{2012}]%
        {mathet2012manual}
\bibfield{author}{\bibinfo{person}{Yann Mathet}, \bibinfo{person}{Antoine
  Widl{\"o}cher}, \bibinfo{person}{Kar{\"e}n Fort}, \bibinfo{person}{Claire
  Fran{\c{c}}ois}, \bibinfo{person}{Olivier Galibert}, \bibinfo{person}{Cyril
  Grouin}, \bibinfo{person}{Juliette Kahn}, \bibinfo{person}{Sophie Rosset},
  {and} \bibinfo{person}{Pierre Zweigenbaum}.} \bibinfo{year}{2012}\natexlab{}.
\newblock \showarticletitle{Manual corpus annotation: Giving meaning to the
  evaluation metrics}. In \bibinfo{booktitle}{\emph{International Conference on
  Computational Linguistics}}. \bibinfo{pages}{809--818}.
\newblock


\bibitem[\protect\citeauthoryear{Mathet, Widl{\"o}cher, and
  M{\'e}tivier}{Mathet et~al\mbox{.}}{2015}]%
        {mathet2015unified}
\bibfield{author}{\bibinfo{person}{Yann Mathet}, \bibinfo{person}{Antoine
  Widl{\"o}cher}, {and} \bibinfo{person}{Jean-Philippe M{\'e}tivier}.}
  \bibinfo{year}{2015}\natexlab{}.
\newblock \showarticletitle{The unified and holistic method gamma ($\gamma$)
  for inter-annotator agreement measure and alignment}.
\newblock \bibinfo{journal}{\emph{Computational Linguistics}}
  \bibinfo{volume}{41}, \bibinfo{number}{3} (\bibinfo{year}{2015}),
  \bibinfo{pages}{437--479}.
\newblock


\bibitem[\protect\citeauthoryear{McClosky, Charniak, and Johnson}{McClosky
  et~al\mbox{.}}{2006}]%
        {mcclosky2006effective}
\bibfield{author}{\bibinfo{person}{David McClosky}, \bibinfo{person}{Eugene
  Charniak}, {and} \bibinfo{person}{Mark Johnson}.}
  \bibinfo{year}{2006}\natexlab{}.
\newblock \showarticletitle{Effective self-training for parsing}. In
  \bibinfo{booktitle}{\emph{Proceedings of the main conference on human
  language technology conference of the North American Chapter of the
  Association of Computational Linguistics}}. Association for Computational
  Linguistics, \bibinfo{pages}{152--159}.
\newblock


\bibitem[\protect\citeauthoryear{Mehar, Matawie, and Maeder}{Mehar
  et~al\mbox{.}}{2013}]%
        {mehar2013determining}
\bibfield{author}{\bibinfo{person}{Arshad~Muhammad Mehar},
  \bibinfo{person}{Kenan Matawie}, {and} \bibinfo{person}{Anthony Maeder}.}
  \bibinfo{year}{2013}\natexlab{}.
\newblock \showarticletitle{Determining an optimal value of K in K-means
  clustering}. In \bibinfo{booktitle}{\emph{2013 IEEE International Conference
  on Bioinformatics and Biomedicine}}. IEEE, \bibinfo{pages}{51--55}.
\newblock


\bibitem[\protect\citeauthoryear{Mehrabi, Morstatter, Saxena, Lerman, and
  Galstyan}{Mehrabi et~al\mbox{.}}{2021}]%
        {mehrabi2021survey}
\bibfield{author}{\bibinfo{person}{Ninareh Mehrabi}, \bibinfo{person}{Fred
  Morstatter}, \bibinfo{person}{Nripsuta Saxena}, \bibinfo{person}{Kristina
  Lerman}, {and} \bibinfo{person}{Aram Galstyan}.}
  \bibinfo{year}{2021}\natexlab{}.
\newblock \showarticletitle{A survey on bias and fairness in machine learning}.
\newblock \bibinfo{journal}{\emph{ACM Computing Surveys (CSUR)}}
  \bibinfo{volume}{54}, \bibinfo{number}{6} (\bibinfo{year}{2021}),
  \bibinfo{pages}{1--35}.
\newblock


\bibitem[\protect\citeauthoryear{Meir, Amir, Cohensius, Ben-Porat, Ben-Shabat,
  and Xia}{Meir et~al\mbox{.}}{2020}]%
        {meir2020truth}
\bibfield{author}{\bibinfo{person}{Reshef Meir}, \bibinfo{person}{Ofra Amir},
  \bibinfo{person}{Gal Cohensius}, \bibinfo{person}{Omer Ben-Porat},
  \bibinfo{person}{Tsviel Ben-Shabat}, {and} \bibinfo{person}{Lirong Xia}.}
  \bibinfo{year}{2020}\natexlab{}.
\newblock \bibinfo{title}{Truth Discovery via Average Proximity}.
\newblock
\newblock
\showeprint[arxiv]{1905.00629}~[cs.AI]


\bibitem[\protect\citeauthoryear{Myers, Well, and Lorch~Jr}{Myers
  et~al\mbox{.}}{2013}]%
        {myers2013research}
\bibfield{author}{\bibinfo{person}{Jerome~L Myers}, \bibinfo{person}{Arnold~D
  Well}, {and} \bibinfo{person}{Robert~F Lorch~Jr}.}
  \bibinfo{year}{2013}\natexlab{}.
\newblock \bibinfo{booktitle}{\emph{Research design and statistical analysis}}.
\newblock \bibinfo{publisher}{Routledge}.
\newblock


\bibitem[\protect\citeauthoryear{Nguyen, Halpern, Wallace, and Lease}{Nguyen
  et~al\mbox{.}}{2016}]%
        {Nguyen16-hcomp}
\bibfield{author}{\bibinfo{person}{An~Thanh Nguyen}, \bibinfo{person}{Matthew
  Halpern}, \bibinfo{person}{Byron~C.\ Wallace}, {and} \bibinfo{person}{Matthew
  Lease}.} \bibinfo{year}{2016}\natexlab{}.
\newblock \showarticletitle{{Probabilistic Modeling for Crowdsourcing
  Partially-Subjective Ratings}}. In \bibinfo{booktitle}{\emph{{Proceedings of
  the 4th AAAI Conference on Human Computation and Crowdsourcing (HCOMP)}}}.
  \bibinfo{pages}{149--158}.
\newblock


\bibitem[\protect\citeauthoryear{Nguyen, Wallace, Li, Nenkova, and
  Lease}{Nguyen et~al\mbox{.}}{2017}]%
        {nguyen2017aggregating}
\bibfield{author}{\bibinfo{person}{An~T Nguyen}, \bibinfo{person}{Byron~C
  Wallace}, \bibinfo{person}{Junyi~Jessy Li}, \bibinfo{person}{Ani Nenkova},
  {and} \bibinfo{person}{Matthew Lease}.} \bibinfo{year}{2017}\natexlab{}.
\newblock \showarticletitle{Aggregating and predicting sequence labels from
  crowd annotations}. In \bibinfo{booktitle}{\emph{Proceedings of the
  conference. Association for Computational Linguistics. Meeting}},
  Vol.~\bibinfo{volume}{2017}. NIH Public Access, \bibinfo{pages}{299}.
\newblock


\bibitem[\protect\citeauthoryear{Northcutt, Athalye, and Mueller}{Northcutt
  et~al\mbox{.}}{2021}]%
        {northcutt2021pervasive}
\bibfield{author}{\bibinfo{person}{Curtis~G Northcutt}, \bibinfo{person}{Anish
  Athalye}, {and} \bibinfo{person}{Jonas Mueller}.}
  \bibinfo{year}{2021}\natexlab{}.
\newblock \showarticletitle{Pervasive Label Errors in Test Sets Destabilize
  Machine Learning Benchmarks}. In \bibinfo{booktitle}{\emph{Thirty-fifth
  Conference on Neural Information Processing Systems: Datasets and Benchmarks
  Track}}.
\newblock


\bibitem[\protect\citeauthoryear{Papineni, Roukos, Ward, and Zhu}{Papineni
  et~al\mbox{.}}{2002}]%
        {papineni2002bleu}
\bibfield{author}{\bibinfo{person}{Kishore Papineni}, \bibinfo{person}{Salim
  Roukos}, \bibinfo{person}{Todd Ward}, {and} \bibinfo{person}{Wei-Jing Zhu}.}
  \bibinfo{year}{2002}\natexlab{}.
\newblock \showarticletitle{Bleu: a method for automatic evaluation of machine
  translation}. In \bibinfo{booktitle}{\emph{Proceedings of the 40th annual
  meeting of the Association for Computational Linguistics}}.
  \bibinfo{pages}{311--318}.
\newblock


\bibitem[\protect\citeauthoryear{Parameswaran, Sarma, and
  Venkataraman}{Parameswaran et~al\mbox{.}}{2016}]%
        {parameswaran2016optimizing}
\bibfield{author}{\bibinfo{person}{Aditya Parameswaran},
  \bibinfo{person}{Akash~Das Sarma}, {and} \bibinfo{person}{Vipul
  Venkataraman}.} \bibinfo{year}{2016}\natexlab{}.
\newblock \showarticletitle{Optimizing open-ended crowdsourcing: The next
  frontier in crowdsourced data management}.
\newblock \bibinfo{journal}{\emph{Bulletin of the Technical Committee on Data
  Engineering}}  \bibinfo{volume}{39} (\bibinfo{year}{2016}).
\newblock


\bibitem[\protect\citeauthoryear{Paun, Artstein, and Poesio}{Paun
  et~al\mbox{.}}{2022}]%
        {paun2022statistical}
\bibfield{author}{\bibinfo{person}{Silviu Paun}, \bibinfo{person}{Ron
  Artstein}, {and} \bibinfo{person}{Massimo Poesio}.}
  \bibinfo{year}{2022}\natexlab{}.
\newblock \showarticletitle{Statistical Methods for Annotation Analysis}.
\newblock \bibinfo{journal}{\emph{Synthesis Lectures on Human Language
  Technologies}} \bibinfo{volume}{15}, \bibinfo{number}{1}
  (\bibinfo{year}{2022}), \bibinfo{pages}{1--217}.
\newblock


\bibitem[\protect\citeauthoryear{Rezatofighi, Tsoi, Gwak, Sadeghian, Reid, and
  Savarese}{Rezatofighi et~al\mbox{.}}{2019}]%
        {rezatofighi2019generalized}
\bibfield{author}{\bibinfo{person}{Hamid Rezatofighi}, \bibinfo{person}{Nathan
  Tsoi}, \bibinfo{person}{JunYoung Gwak}, \bibinfo{person}{Amir Sadeghian},
  \bibinfo{person}{Ian Reid}, {and} \bibinfo{person}{Silvio Savarese}.}
  \bibinfo{year}{2019}\natexlab{}.
\newblock \showarticletitle{Generalized intersection over union: A metric and a
  loss for bounding box regression}. In \bibinfo{booktitle}{\emph{Proceedings
  of the IEEE/CVF Conference on Computer Vision and Pattern Recognition}}.
  \bibinfo{pages}{658--666}.
\newblock


\bibitem[\protect\citeauthoryear{Ruggero~Ronchi and Perona}{Ruggero~Ronchi and
  Perona}{2017}]%
        {ruggero2017benchmarking}
\bibfield{author}{\bibinfo{person}{Matteo Ruggero~Ronchi} {and}
  \bibinfo{person}{Pietro Perona}.} \bibinfo{year}{2017}\natexlab{}.
\newblock \showarticletitle{Benchmarking and error diagnosis in multi-instance
  pose estimation}. In \bibinfo{booktitle}{\emph{Proceedings of the IEEE
  international conference on computer vision}}. \bibinfo{pages}{369--378}.
\newblock


\bibitem[\protect\citeauthoryear{Sang and De~Meulder}{Sang and
  De~Meulder}{2003}]%
        {sang2003introduction}
\bibfield{author}{\bibinfo{person}{Erik Tjong~Kim Sang} {and}
  \bibinfo{person}{Fien De~Meulder}.} \bibinfo{year}{2003}\natexlab{}.
\newblock \showarticletitle{Introduction to the CoNLL-2003 Shared Task:
  Language-Independent Named Entity Recognition}. In
  \bibinfo{booktitle}{\emph{Proceedings of the Seventh Conference on Natural
  Language Learning at HLT-NAACL 2003}}. \bibinfo{pages}{142--147}.
\newblock


\bibitem[\protect\citeauthoryear{Scholer, Kelly, Wu, Lee, and Webber}{Scholer
  et~al\mbox{.}}{2013}]%
        {scholer2013effect}
\bibfield{author}{\bibinfo{person}{Falk Scholer}, \bibinfo{person}{Diane
  Kelly}, \bibinfo{person}{Wan-Ching Wu}, \bibinfo{person}{Hanseul~S Lee},
  {and} \bibinfo{person}{William Webber}.} \bibinfo{year}{2013}\natexlab{}.
\newblock \showarticletitle{The effect of threshold priming and need for
  cognition on relevance calibration and assessment}. In
  \bibinfo{booktitle}{\emph{Proceedings of the 36th international ACM SIGIR
  conference on Research and development in information retrieval}}.
  \bibinfo{pages}{623--632}.
\newblock


\bibitem[\protect\citeauthoryear{Scott}{Scott}{1955}]%
        {scott1955reliability}
\bibfield{author}{\bibinfo{person}{William~A Scott}.}
  \bibinfo{year}{1955}\natexlab{}.
\newblock \showarticletitle{Reliability of content analysis: The case of
  nominal scale coding}.
\newblock \bibinfo{journal}{\emph{Public opinion quarterly}}
  (\bibinfo{year}{1955}), \bibinfo{pages}{321--325}.
\newblock


\bibitem[\protect\citeauthoryear{Sekine and Collins}{Sekine and
  Collins}{1997}]%
        {sekine1997evalb}
\bibfield{author}{\bibinfo{person}{Satoshi Sekine} {and}
  \bibinfo{person}{Michael Collins}.} \bibinfo{year}{1997}\natexlab{}.
\newblock \bibinfo{title}{EvalB: a bracket scoring program}.
\newblock
\newblock
\urldef\tempurl%
\url{http://nlp.cs.nyu.edu/evalb/}
\showURL{%
\tempurl}


\bibitem[\protect\citeauthoryear{Sen, Giesel, Gold, Hillmann, Lesicko, Naden,
  Russell, Wang, and Hecht}{Sen et~al\mbox{.}}{2015}]%
        {sen2015turkers}
\bibfield{author}{\bibinfo{person}{Shilad Sen}, \bibinfo{person}{Margaret~E
  Giesel}, \bibinfo{person}{Rebecca Gold}, \bibinfo{person}{Benjamin Hillmann},
  \bibinfo{person}{Matt Lesicko}, \bibinfo{person}{Samuel Naden},
  \bibinfo{person}{Jesse Russell}, \bibinfo{person}{Zixiao Wang}, {and}
  \bibinfo{person}{Brent Hecht}.} \bibinfo{year}{2015}\natexlab{}.
\newblock \showarticletitle{Turkers, scholars," arafat" and" peace" cultural
  communities and algorithmic gold standards}. In
  \bibinfo{booktitle}{\emph{Proceedings of the 18th acm conference on computer
  supported cooperative work \& social computing}}. \bibinfo{pages}{826--838}.
\newblock


\bibitem[\protect\citeauthoryear{Skj{\ae}rholt}{Skj{\ae}rholt}{2014}]%
        {skjaerholt2014chance}
\bibfield{author}{\bibinfo{person}{Arne Skj{\ae}rholt}.}
  \bibinfo{year}{2014}\natexlab{}.
\newblock \showarticletitle{A chance-corrected measure of inter-annotator
  agreement for syntax}. In \bibinfo{booktitle}{\emph{Proceedings of the 52nd
  Annual Meeting of the Association for Computational Linguistics (Volume 1:
  Long Papers)}}. \bibinfo{pages}{934--944}.
\newblock


\bibitem[\protect\citeauthoryear{Snow, O'Connor, Jurafsky, and Ng}{Snow
  et~al\mbox{.}}{2008}]%
        {snow2008cheap}
\bibfield{author}{\bibinfo{person}{Rion Snow}, \bibinfo{person}{Brendan
  O'Connor}, \bibinfo{person}{Daniel Jurafsky}, {and} \bibinfo{person}{Andrew~Y
  Ng}.} \bibinfo{year}{2008}\natexlab{}.
\newblock \showarticletitle{Cheap and fast---but is it good?: evaluating
  non-expert annotations for natural language tasks}. In
  \bibinfo{booktitle}{\emph{Proceedings of the conference on empirical methods
  in natural language processing}}. Association for Computational Linguistics,
  \bibinfo{pages}{254--263}.
\newblock


\bibitem[\protect\citeauthoryear{Tian and Zhu}{Tian and Zhu}{2012}]%
        {tian2012learning}
\bibfield{author}{\bibinfo{person}{Yuandong Tian} {and} \bibinfo{person}{Jun
  Zhu}.} \bibinfo{year}{2012}\natexlab{}.
\newblock \showarticletitle{Learning from crowds in the presence of schools of
  thought}. In \bibinfo{booktitle}{\emph{Proceedings of the 18th ACM SIGKDD
  international conference on Knowledge discovery and data mining}}.
  \bibinfo{pages}{226--234}.
\newblock


\bibitem[\protect\citeauthoryear{Whitehill, Wu, Bergsma, Movellan, and
  Ruvolo}{Whitehill et~al\mbox{.}}{2009}]%
        {whitehill2009whose}
\bibfield{author}{\bibinfo{person}{Jacob Whitehill}, \bibinfo{person}{Ting-fan
  Wu}, \bibinfo{person}{Jacob Bergsma}, \bibinfo{person}{Javier Movellan},
  {and} \bibinfo{person}{Paul Ruvolo}.} \bibinfo{year}{2009}\natexlab{}.
\newblock \showarticletitle{Whose vote should count more: Optimal integration
  of labels from labelers of unknown expertise}.
\newblock \bibinfo{journal}{\emph{Advances in neural information processing
  systems}}  \bibinfo{volume}{22} (\bibinfo{year}{2009}).
\newblock


\bibitem[\protect\citeauthoryear{Wu, Schuster, Chen, Le, Norouzi, Macherey,
  Krikun, Cao, Gao, Macherey, et~al\mbox{.}}{Wu et~al\mbox{.}}{2016}]%
        {wu2016google}
\bibfield{author}{\bibinfo{person}{Yonghui Wu}, \bibinfo{person}{Mike
  Schuster}, \bibinfo{person}{Zhifeng Chen}, \bibinfo{person}{Quoc~V Le},
  \bibinfo{person}{Mohammad Norouzi}, \bibinfo{person}{Wolfgang Macherey},
  \bibinfo{person}{Maxim Krikun}, \bibinfo{person}{Yuan Cao},
  \bibinfo{person}{Qin Gao}, \bibinfo{person}{Klaus Macherey}, {et~al\mbox{.}}}
  \bibinfo{year}{2016}\natexlab{}.
\newblock \showarticletitle{Google's neural machine translation system:
  Bridging the gap between human and machine translation}.
\newblock \bibinfo{journal}{\emph{arXiv preprint arXiv:1609.08144}}
  (\bibinfo{year}{2016}).
\newblock


\bibitem[\protect\citeauthoryear{Yang and Jin}{Yang and Jin}{2006}]%
        {yang2006distance}
\bibfield{author}{\bibinfo{person}{Liu Yang} {and} \bibinfo{person}{Rong Jin}.}
  \bibinfo{year}{2006}\natexlab{}.
\newblock \showarticletitle{Distance metric learning: A comprehensive survey}.
\newblock \bibinfo{journal}{\emph{Michigan State Universiy}}
  \bibinfo{volume}{2}, \bibinfo{number}{2} (\bibinfo{year}{2006}),
  \bibinfo{pages}{4}.
\newblock


\bibitem[\protect\citeauthoryear{Zhang, Kishore, Wu, Weinberger, and
  Artzi}{Zhang et~al\mbox{.}}{2019}]%
        {zhang2019bertscore}
\bibfield{author}{\bibinfo{person}{Tianyi Zhang}, \bibinfo{person}{Varsha
  Kishore}, \bibinfo{person}{Felix Wu}, \bibinfo{person}{Kilian~Q Weinberger},
  {and} \bibinfo{person}{Yoav Artzi}.} \bibinfo{year}{2019}\natexlab{}.
\newblock \showarticletitle{Bertscore: Evaluating text generation with bert}.
\newblock \bibinfo{journal}{\emph{arXiv preprint arXiv:1904.09675}}
  (\bibinfo{year}{2019}).
\newblock


\end{thebibliography}

\clearpage 

\section{Appendix}

\textbf{Distance functions}

\textbf{Vectors: Binary}
$$
D(a, b) = 1 - \frac{1}{N} \sum_i^N 1_{a_i=b_i}
$$

\textbf{Vectors: Euclidean}
$$
D(a, b) = 1 - \textrm{RMSE}(a, b)
$$
$$
\textrm{RMSE}(a, b)= \sqrt{ \frac{1}{N} \sum_i^N (a_i-b_i)^2 }
$$

\textbf{Translations: Levenshtein}

\url{https://en.wikipedia.org/wiki/Levenshtein_distance}

(on tokens not characters)

\textbf{Translations: BLEU}

$$
D(a, b) = \frac{\textrm{bleu}(a, b) +  \textrm{bleu}(b, a)}{2}
$$

bleu: nltk version 3.4.5 sentence\_bleu
smoothing method 4

\textbf{Translations: GLEU}

$$
D(a, b) = \frac{\textrm{gleu}(a, b) +  \textrm{gleu}(b, a)}{2}
$$

gleu: nltk version 3.4.5 sentence\_gleu
smoothing method 4

\textbf{Translations: BERTScore}

$D(a, b) = 1 - F1$ using BERTScorer.score(a, b) from bert\_score version 0.3.10

\textbf{Various: Count Diff}
$$
D_m(A, B) = ||A| - |B||
$$

\textbf{Bounding Box and Keypoints: single to multi}
$$
D_m(A,B) = \frac{\Delta(A,B) + \Delta(B,A)}{2}
$$
$$
\Delta(A,B) = \mathbb{E} \{ \text{min}(\{D_s(a, b) \mid b \in B \}) \mid a \in A \} \\
$$

\textbf{Bounding Box: L2}
$$
D_s(a, b) = 1 - \frac{\textrm{RMSE}(a_0, b_0) + \textrm{RMSE}(a_F, b_F)}{20}
$$

\textbf{Bounding Box: IoU Score}
$$
D_s(a, b) = \frac{\cap(a, b)}{\textrm{AREA}(a) + \textrm{AREA}(b) - \cap(a, b)}
$$

\textbf{Bounding Box: GIoU Score}
Adjustment to IoU described in \citet{rezatofighi2019generalized}.

\textbf{Keypoints: OKS Score}
OKS distance function described in \citet{ruggero2017benchmarking}.

\textbf{NER: single to multi}
$$
D_m(A,B) = 1 - \frac{2 \Delta(A,B) \Delta(B,A)}{\Delta(A,B) + \Delta(B,A)}
$$

\textbf{NER: both lenient}
$$
\Delta(A,B) = \mathbb{E} \{ \frac{\sum_{t \in a}\sum_{s \in b \mid b \in B}1_{s=t}}{|\{t \in a\}|} \mid a \in A \} \\
$$

\textbf{NER: strict tag}
$$
\Delta(A,B) = \mathbb{E} \{ \frac{\sum_{t \in a}\sum_{s \in b \mid b \in B}1_{s=t \land \textrm{TAG}(a)=\textrm{TAG}(b)}}{|\{t \in a\}|} \mid a \in A \} \\
$$

\textbf{NER: strict range}
$$
\Delta(A,B) = \frac{\sum_{a \in A}\sum_{b \in B}1_{a=b}}{|A|} \\
$$

\textbf{NER: both strict}
$$
\Delta(A,B) = \frac{\sum_{a \in A}\sum_{b \in B}1_{a=b \land \textrm{TAG}(a)=\textrm{TAG}(b)}}{|A|} \\
$$

\textbf{Parse Trees: $\alpha_{\textrm{plain}}$}
$$
D(a,b) = \textrm{TED}(a, b)
$$

TED: zss version 1.2.0 simple\_distance

\textbf{Parse Trees: $\alpha_{\textrm{diff}}$}
$$
D(a,b) = \textrm{TED}(a, b) - |\textrm{NLEAVES(a)} - \textrm{NLEAVES(b)}|
$$

\textbf{Parse Trees: $\alpha_{\textrm{norm}}$}
$$
D(a,b) = \frac{\textrm{TED}(a, b)}{\textrm{NLEAVES(a)} + \textrm{NLEAVES(b)}}
$$

\textbf{Ranked Lists: Kendall's $\tau$}

scipy version 1.3.1 stats.kendalltau

\textbf{Ranked Lists: Spearmans' $\rho$}

scipy version 1.3.1 stats.spearmanr

\vspace{2.5em} 

\begin{table}[h]
\centering
\begin{tabular}{llrr}
\toprule
Dataset & Annotators & Items & Annotations \\
\midrule
Vectors  & 38 & 100 & 1000  \\
Translations  & 70 & 250 & 2490 \\
Bounding Box  & 196 & 200 & 1723 \\
NER  & 46 & 199 & 982 \\
Keypoints   & 100 & 199 & 1000  \\
Parse Trees  & 24 & 128 & 512  \\
Ranked Lists   & 30 & 100 & 600  \\
\bottomrule
\end{tabular}
\caption{Datasets used and summary statistics}
\end{table}

\begin{figure}[h]
\centering
\includegraphics[clip,width=\columnwidth]{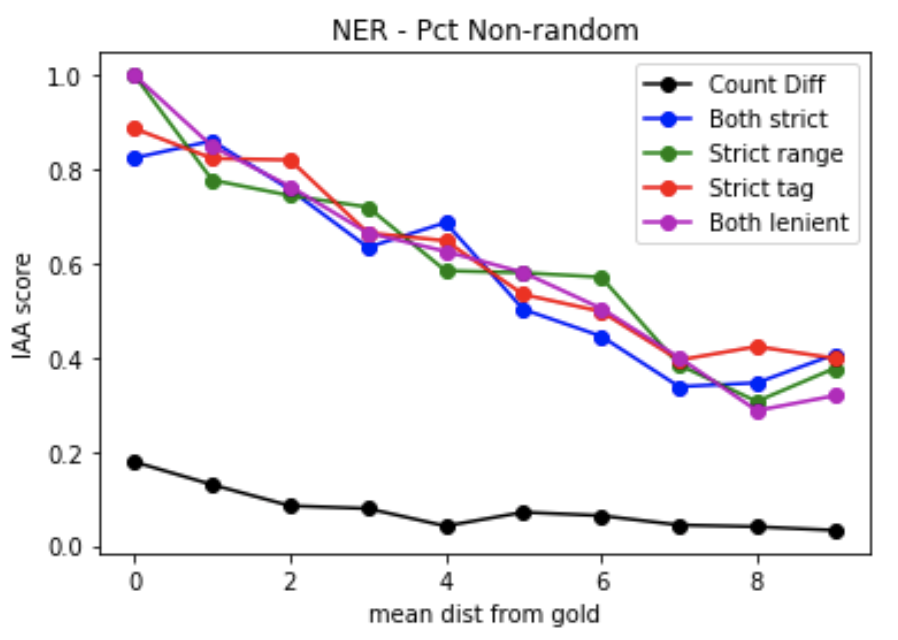}
\caption{Unsurprisingly, $\sigma$ measure decreases with increased noise (expected distance from gold), NER example.}
\label{fig:ner-cst}
\end{figure}

\clearpage

\begin{figure}[!h]
  \centering
  
  \begin{subfigure}{.55\columnwidth}
    \centering
    \includegraphics[width=\linewidth]{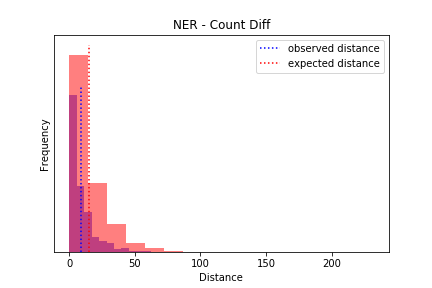}
  \end{subfigure}%
  \begin{subfigure}{.55\columnwidth}
    \centering
    \includegraphics[width=\linewidth]{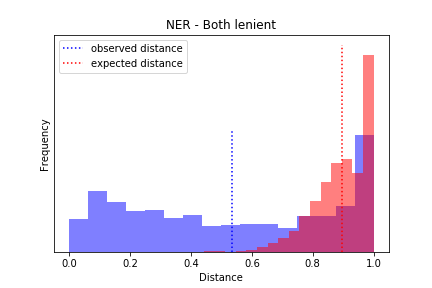}
  \end{subfigure} \\
  \begin{subfigure}{.55\columnwidth}
    \centering
    \includegraphics[width=\linewidth]{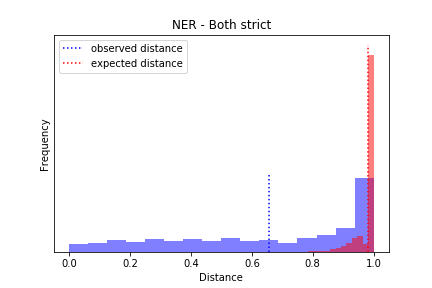}
  \end{subfigure}%
  \begin{subfigure}{.55\columnwidth}
    \centering
    \includegraphics[width=\linewidth]{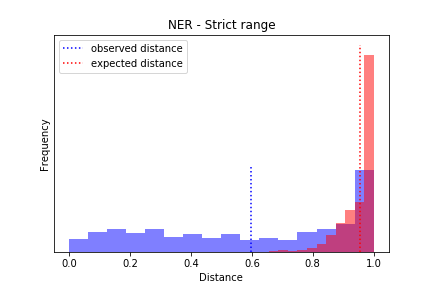}
  \end{subfigure} \\
  \begin{subfigure}{.55\columnwidth}
    \centering
    \includegraphics[width=\linewidth]{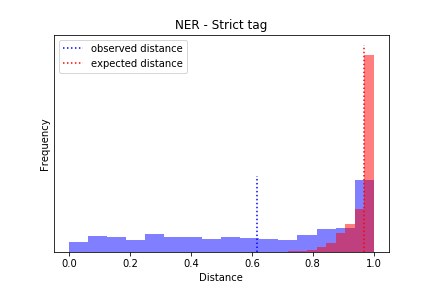}
  \end{subfigure}%
  \begin{subfigure}{.55\columnwidth}
    \centering
    \includegraphics[width=\linewidth]{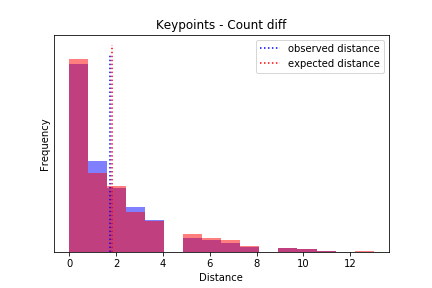}
  \end{subfigure} \\
  \begin{subfigure}{.55\columnwidth}
    \centering
    \includegraphics[width=\linewidth]{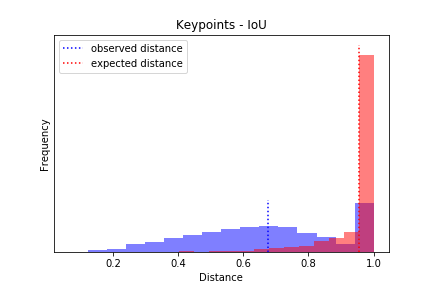}
  \end{subfigure}%
  \begin{subfigure}{.55\columnwidth}
    \centering
    \includegraphics[width=\linewidth]{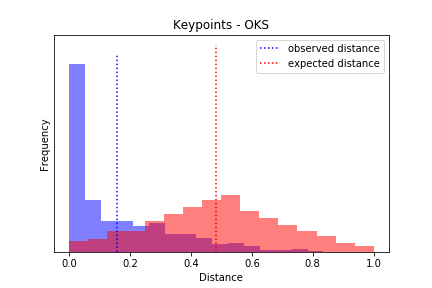}
  \end{subfigure} \\
  \begin{subfigure}{.55\columnwidth}
    \centering
    \includegraphics[width=\linewidth]{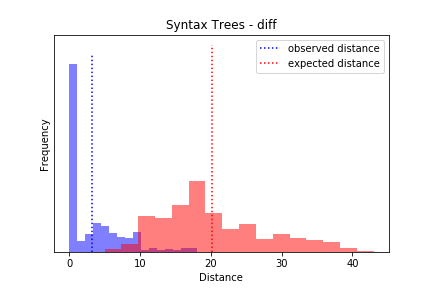}
  \end{subfigure}%
  \begin{subfigure}{.55\columnwidth}
    \centering
    \includegraphics[width=\linewidth]{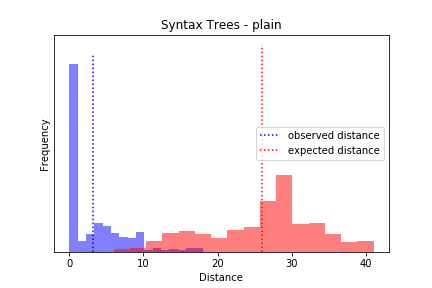}
  \end{subfigure} \\
  \begin{subfigure}{.55\columnwidth}
    \centering
    \includegraphics[width=\linewidth]{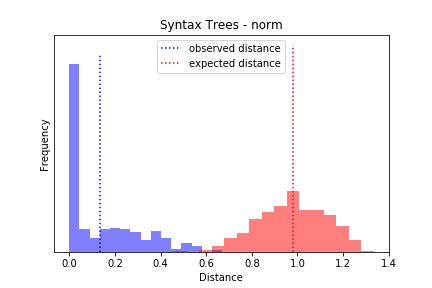}
  \end{subfigure}%
  \caption{Distributions of expected and observed distances across all datasets and distance functions.}
\end{figure} 
\begin{figure}[!h]
  \centering
  
  \begin{subfigure}{.55\columnwidth}
    \centering
    \includegraphics[width=\linewidth]{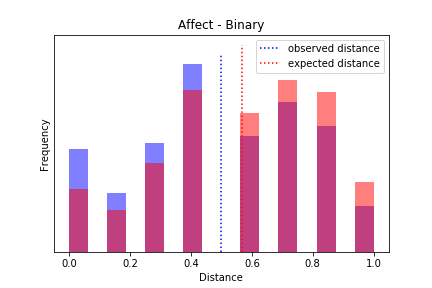}
  \end{subfigure}%
  \begin{subfigure}{.55\columnwidth}
    \centering
    \includegraphics[width=\linewidth]{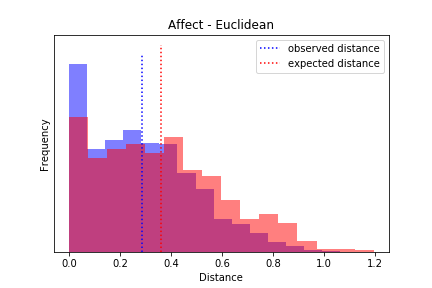}
  \end{subfigure} \\
  \begin{subfigure}{.55\columnwidth}
    \centering
    \includegraphics[width=\linewidth]{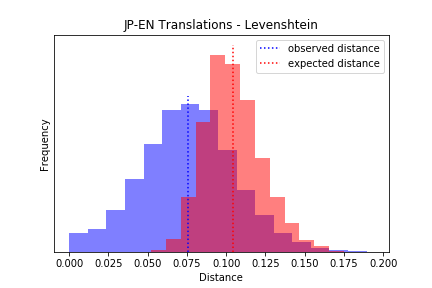}
  \end{subfigure}%
  \begin{subfigure}{.55\columnwidth}
    \centering
    \includegraphics[width=\linewidth]{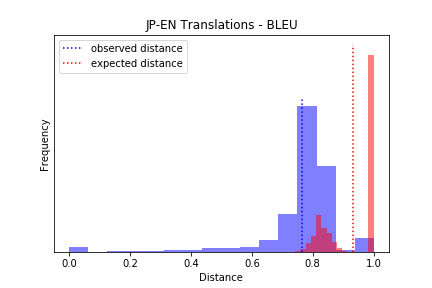}
  \end{subfigure} \\
  \begin{subfigure}{.55\columnwidth}
    \centering
    \includegraphics[width=\linewidth]{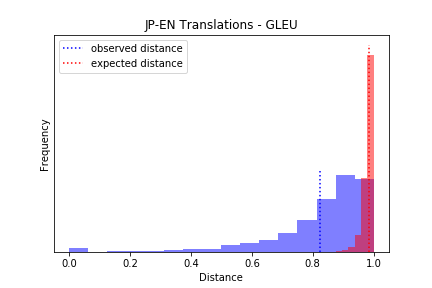}
  \end{subfigure}%
  \begin{subfigure}{.55\columnwidth}
    \centering
    \includegraphics[width=\linewidth]{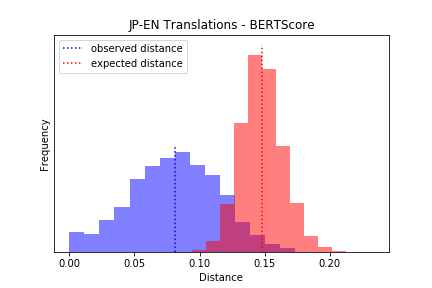}
  \end{subfigure} \\
  \begin{subfigure}{.55\columnwidth}
    \centering
    \includegraphics[width=\linewidth]{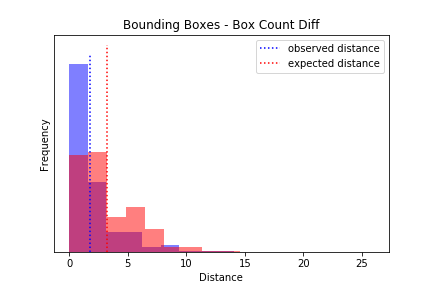}
  \end{subfigure}%
  \begin{subfigure}{.55\columnwidth}
    \centering
    \includegraphics[width=\linewidth]{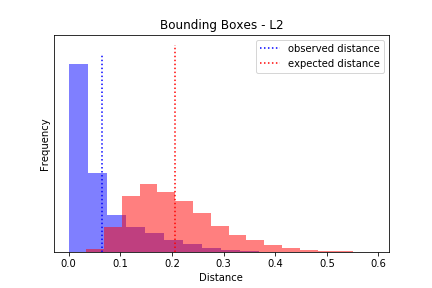}
  \end{subfigure} \\
  \begin{subfigure}{.55\columnwidth}
    \centering
    \includegraphics[width=\linewidth]{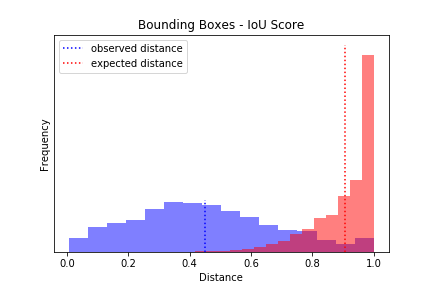}
  \end{subfigure}%
  \begin{subfigure}{.55\columnwidth}
    \centering
    \includegraphics[width=\linewidth]{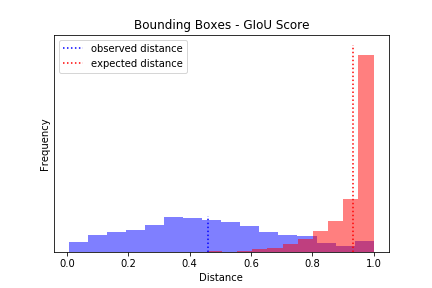}
  \end{subfigure} \\
  \begin{subfigure}{.55\columnwidth}
    \centering
    \includegraphics[width=\linewidth]{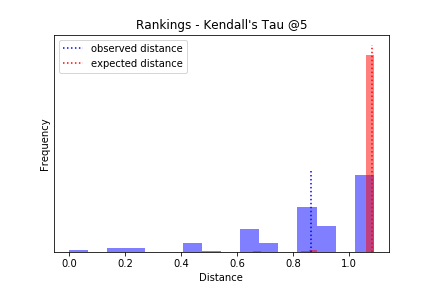}
  \end{subfigure}%
  \begin{subfigure}{.55\columnwidth}
    \centering
    \includegraphics[width=\linewidth]{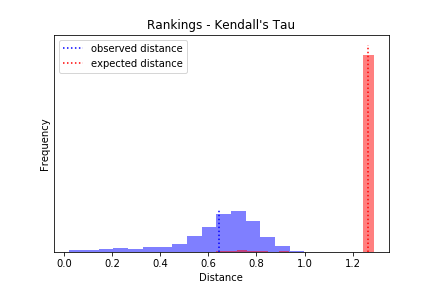}
  \end{subfigure} \\
  \begin{subfigure}{.55\columnwidth}
    \centering
    \includegraphics[width=\linewidth]{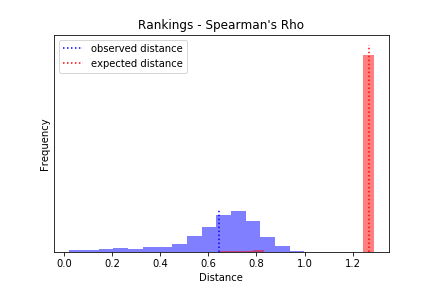}
  \end{subfigure}%
  
\end{figure}

\end{document}